\documentclass{article} 
\usepackage{collas2024_conference,times}
\usepackage{easyReview}

\usepackage{amsmath,amsfonts,bm}

\def\eqref#1{equation~\ref{#1}}

\def\1{\bm{1}}

\def\ra{{\textnormal{a}}}

\def\mI{{\bm{I}}}

\DeclareMathAlphabet{\mathsfit}{\encodingdefault}{\sfdefault}{m}{sl}
\SetMathAlphabet{\mathsfit}{bold}{\encodingdefault}{\sfdefault}{bx}{n}

\usepackage{hyperref}
\hypersetup{
    colorlinks=true,
    linkcolor=red,
    filecolor=magenta,
    urlcolor=blue,
    citecolor=purple,
    pdftitle={Overleaf Example},
    pdfpagemode=FullScreen,
    }
\usepackage{amsmath,amsfonts,amssymb, amsthm} 
\usepackage[capitalize,noabbrev]{cleveref}

\usepackage{bbold}
\usepackage{multirow}
\usepackage{bm}
\usepackage{booktabs}
\usepackage{wrapfig}
\usepackage{authblk}

\theoremstyle{plain}
\newtheorem{theorem}{Theorem}[section]

\theoremstyle{definition}
\newtheorem{definition}[theorem]{Definition}
\newtheorem{assumption}[theorem]{Assumption}
\theoremstyle{remark}

\title{Local vs Global continual learning}

\author[1,2]{Giulia Lanzillotta}
\author[2]{Sidak Pal Singh}
\author[3]{Benjamin F. Grewe}
\author[2]{Thomas Hofmann}
\affil[1]{ETH AI Center, Switzerland}
\affil[2]{Department of Computer Science, ETH Zurich, Switzerland}
\affil[3]{Institute of Neuroinformatics - University of Zurich and ETH Zurich, Switzerland}
\affil[ ]{\textit{Correspondence address:  giulia.lanzillotta@ai.ethz.ch}}

\collasfinalcopy 

\begin{document}

\maketitle

\begin{abstract}
Continual learning is the problem of  integrating new information in a model while retaining the knowledge acquired in the past. 
Despite the tangible improvements achieved in recent years, {the problem of continual learning is still an open one}. 
A better understanding of the mechanisms behind the successes and failures of existing continual learning algorithms can unlock the development of new successful strategies. 
In this work, we view continual learning from the perspective of the \emph{multi-task loss approximation}, and we compare two alternative strategies, namely \emph{local} and \emph{global} approximations. We classify existing continual learning algorithms based on the approximation used, and we assess the practical effects of this distinction in common continual learning settings. Additionally, we study optimal continual learning objectives in the case of local polynomial approximations and we provide examples of existing algorithms implementing the optimal objectives.

\end{abstract}

\section{Introduction}
 
Given the present trend toward training gigantic, foundational models~\citep{achiam2023gpt}, effective continual learning solutions hold the key to reducing the computational costs of ever integrating new information into the model without forgetting older information. The alternative of retraining on the entire data to update the model is not an option at this scale, especially in the case where the models are expected to adapt quickly to a dynamic environment. 

\citet{mccloskey1989catastrophic} were the first to observe that neural networks trained on a sequence of tasks perform poorly on inputs from temporally antecedent tasks, a phenomenon termed \emph{catastrophic forgetting}. Several algorithms have since been developed to limit catastrophic forgetting in deep neural networks \citep{de2021continual,khetarpal2022towards}. {However the solutions developed so far struggle in real world scenarios, where satisfying either compute or memory constraints is crucial \citep{kontogianni2024continual, garg2023tic}}. We believe that a better understanding of the problem of continual learning is needed to design algorithms that can address catastrophic forgetting effectively. Moreover, studying the failure cases of existing continual learning algorithms can point the way towards principled solutions.

In this work, we view continual learning as the problem of \emph{approximating the multi-task loss} and we study existing continual learning algorithms in this light. In particular, we are interested in distinguishing between \emph{local} and \emph{global} approximations. Said simply, local approximations rely on information about the state of the network after learning each task, whereas global approximations don't. This characteristic implies that algorithms employing local approximations need to satisfy a restrictive assumption, which we bring to the surface. 
We classify continual learning algorithms as either \emph{local} or \emph{global}, based on the properties of the underlying approximation, and we evaluate experimentally the practical consequences of the two approximation schemes. 

\textbf{Contributions and paper structure.} After covering some background (\cref{background-continual}) we introduce the formulation of continual learning from the perspective of loss approximation (\cref{problem-form}). In \cref{lgcldefinition} we discuss two mutually exclusive approximation strategies, namely \emph{local} and \emph{global} approximations and we define the locality assumption, which is unavoidable for any algorithm employing a local approximation. Next (\cref{local-poly-apprx}), we study the case of continual learning algorithms using polynomial local approximations, and we derive optimal objectives for the quadratic case. In \cref{literature-examples} we consider a few classic examples from the literature and we classify them as either local or global algorithms. Further, we show formally that orthogonal gradient descent \citep{farajtabar2020orthogonal} implements the optimal objective for local quadratic approximations derived in \cref{local-poly-apprx}. Finally, we provide experimental evidence of our claims in \cref{experiments}, and we evaluate the practical implications of local and global approximations in common continual learning settings.

\section{Background}

\label{background-continual}

Consider a set of supervised learning tasks $\mathcal{T}_1, \dots, \mathcal{T}_T$, with the data associated to the $t$-th task $\mathcal{T}_t$ being ${D}_t = \{(x,y) \in \mathcal{X}_t\times \mathcal{Y}_t\}$. Continual learning algorithms learn the tasks \emph{sequentially}, whereby at each \emph{learning step} they utilize the present task data 
to update the model, which is typically a neural network with parameters $\bm\theta$. In order to avoid forgetting the old tasks while learning a new one, the algorithm can often access an external memory, which is also updated after every task. The content of this memory may differ across algorithms. The performance on each task is measured by a task-specific loss function $l_t$, and the objective of each task is to minimise the expected loss: $\mathcal{L}_t(\bm{\theta}) = \langle l_t(x,y,\bm{\theta})\rangle_{\mathcal{P}_t}$, where $\langle \cdot \rangle_{\mathcal{P}_t}$ denotes the average over the task distribution $\mathcal{P}_t$. In practice, the expectation is approximated by an average over the dataset, which is called the task empirical loss ${{L}}_t(\bm{\theta}) = \langle l_t(x,y,\bm{\theta})\rangle_{D_t}$. The continual learning problem is to minimise the \emph{multi-task loss}: 
\begin{equation}
\label{multi-task-expected}
    \mathcal{L}^{MT}_t(\bm\theta) = \frac{1}{t} \left( {\mathcal{L}}_1(\bm{\theta}) + \dots + {\mathcal{L}}_t(\bm{\theta})\right) 
\end{equation}
{\textit{while only having direct access to the current task data $D_t$ and the external memory}.}
{Other objectives may be considered instead of the multi-task loss, such as the average lifetime performance. We choose to use the multi-task loss objective in order to conform with the historically prevalent practice in the literature.}

\textbf{Notation \& Metrics.}
We use $\bm\theta \in \Theta_t$ to refer to a generic vector in the parameter space and $\bm\theta_t$ to the value of the network parameters after $t$ learning steps. We index the current task with $t$ and any {single} old task with $o$. 
Given a current parameter vector $\bm\theta$ \emph{catastrophic forgetting} on task $\mathcal{T}_o$ may be measured by the signed difference in expected loss 
$\mathcal{E}_o(\bm\theta) = \mathcal{L}_o(\bm\theta) - \mathcal{L}_o(\bm\theta_o)$.
Similarly, the empirical approximation of $\mathcal{E}_o(\bm\theta)$ is the signed change in the empirical task loss $E_o(\bm\theta) = {L}_o(\bm\theta) - {L}_o(\bm\theta_o)$. Alternatively, forgetting can be measured in terms of the task test accuracy as $\mathcal{E}^{acc}_o(\bm\theta) = \operatorname{ACC}_o(\bm\theta_o) - \operatorname{ACC}_o(\bm\theta)$. Notice that in both cases a lower value of forgetting is better. Additionally, we denote the \emph{average forgetting} after $t$ steps by $E(t)= \frac{1}{t} \sum_{o=1}^{t} \mathcal{E}_o(\bm\theta_t)$ and the \emph{average empirical forgetting} as $E(t)$.
Likewise, the \emph{average accuracy} after $t$ steps by $\operatorname{ACC}(t) = \frac{1}{t} \sum_{o=1}^{t} \operatorname{ACC}_o(\bm\theta_t)$. We refer the reader to \cref{AA} (Appendix A) for a full overview of the notation used.

\section{Local and Global approximations in continual learning}
\label{lgcl}

\subsection{Problem formulation}
\label{problem-form}

In this work, we view continual learning algorithms from the perspective of \emph{loss approximation}, which we introduce in this section. Consider the problem of minimizing the multi-task expected loss in \cref{multi-task-expected}. If hypothetically all the data were available, one could approximate the distribution multi-task loss $\mathcal{L}_t^{MT}(\bm\theta)$ by its \emph{empirical version} $L_t^{MT}(\bm\theta)$ and use it as the optimization objective, as is common practice in machine learning. However, a continual learning algorithm can only access the current task data $D_t$ and the content of its external memory $M_t$, therefore, its objective should be, strictly speaking, a function of $D_t$ and $M_t$ alone: $\operatorname{obj}_t(D_t,M_t)$. 

In this work, we are interested in the way in which the (explicit or implicit) objective of a continual learning algorithm $\operatorname{obj}_t(D_t,M_t)$ approximates the true objective, i.e. the multi-task loss. In particular, we consider the relation between $L_t^{MT}(\bm\theta)$ and $\operatorname{obj}_t(D_t,M_t)$ in order to focus on the locality (or non-locality) of the approximation, and we set aside a discussion of the generalization gap (i.e., whether $L_t^{MT}(\bm\theta) \approx \mathcal{L}_t^{MT}(\bm\theta))$ . 
Accordingly, we view  the continual learning problem as follows: 
\begin{equation}
\begin{aligned}
\label{cl-formulation}
&\min_{\bm\theta \in \Theta} \,\, \operatorname{obj}_t(D_t,M_t)(\bm\theta)
    & \,\,\,\,\, s.t. \,\,\, \operatorname{obj}_t(D_t,M_t)(\bm\theta) \approx L_t^{MT}(\bm\theta)
\end{aligned}
\end{equation}
To conform with existing algorithms, we hereafter study approximate objectives $\operatorname{obj}_t(D_t,M_t)$ of the form: 
\begin{equation}
\label{approxmtloss}
    \operatorname{obj}_t(D_t,M_t) = \hat{L}^{MT}_{t}(\bm\theta) := \frac{1}{t} \left( {\hat{L}}_1(\bm{\theta}) + \dots + {\hat{L}}_{t-1}(\bm{\theta}) + {{L}}_t(\bm{\theta})\right) 
\end{equation}
In short, each previous task loss is approximated separately as $\hat{L}_i(\bm{\theta}) \approx {L}_i(\bm{\theta})$. We want to  stress that such an approximation $\hat{L}_i(\bm{\theta})$ need not always be explicit. For some cases, we infer the approximation used from the objective which the algorithm effectively minimizes at each learning step. This distinction will become clear in \cref{literature-examples}, where we review examples from the literature.

\subsection{Local and global approximations}
\label{lgcldefinition}

We are now ready to introduce the central point of our discussion. Starting from the formulation of continual learning introduced in the previous section (\cref{cl-formulation}), we are interested in asking whether the task loss approximation $\hat{L}_t(\bm\theta)$ is \emph{local} or \emph{global}.

In general, a \emph{local} approximation of a function $f(x)$ makes use of information about the function at a particular point $x_0$ to produce a good approximation of $f$ in a neighborhood of $x_0$. 
Correspondingly, we say that the task loss approximation $\hat{L}_t(\bm\theta)$ is \emph{local} when it uses information about the function at the task solution $\bm\theta_t$ and it is accurate in a neighborhood of $\bm\theta_t$. Conversely, we say that the approximation is \emph{global} when it is independent of $\bm\theta_t$, meaning that modifying $\bm\theta_t$ would not change the approximation. 
{ We formalise the notion of local and global approximations in \cref{local-and-global-approx-definition}. 
\begin{definition}[Local and global task loss approximation.] 
\label{local-and-global-approx-definition} Let $I(X;Y)$ denote the mutual information of the pair of random variables $(X,Y)$. We say that the task loss approximation $\hat{L}_t(\bm\theta)$ is \emph{local} when $I(\hat{L}_t(\bm\theta); \bm\theta_t) > 0 \,\,\forall \,\,\bm\theta \in \Theta$, and that it is \emph{global} when $I(\hat{L}_t(\bm\theta); \bm\theta_t) = 0 \,\,\forall \,\,\bm\theta \in \Theta\setminus\{\bm\theta_t\}$.
\end{definition}
}

For illustrative purposes, consider the constant function approximation $\hat{L}_t(\bm\theta) = C$. A local approximation could be $\hat{L}_t(\bm\theta) = L_t(\bm\theta_t)$, and a global approximation could be $\hat{L}_t(\bm\theta) = 0$. {Although it is intuitively clear that the former approximation relies on information of the task solution $\bm\theta_t$, in general one could evaluate $\mathcal{D}_{KL}(\mathcal{P}(L_t(\bm\theta_t), \bm\theta_t) \| \mathcal{P}(L_t(\bm\theta_t))\cdot \mathcal{P}(\bm\theta_t))>0$, given any distribution $\mathcal{P}(\bm\theta_t)$}.

Let $\xi_t(\bm\theta) = |\hat{L}_t(\bm\theta) - L_t(\bm\theta)|$ be the approximation error in $\bm\theta$. We define the \emph{$\epsilon$-region} of the approximation as  $\Omega^\epsilon_t = \{\bm\theta \in \Theta : \xi_t(\bm\theta) < \epsilon\}$. For local approximations $\Omega^\epsilon_t$ is always a neighborhood of $\bm\theta_t$  (this may be another definition of local approximations), while it may be a disjoint set of points for a global approximation. Notice that the global approximation is not necessarily more accurate than a local approximation. Depending on the case, the $\epsilon$-region of a local approximation may have more volume than the $\epsilon$-region of a global approximation, and thus be `less wrong' on average. 

This brings us to state the main assumption for continual learning algorithms using local approximations. Hereafter, we say a continual learning algorithm is \emph{local} if it uses a local approximation of the task loss function, and \emph{global} if it uses a global approximation instead. 
It follows that local continual learning algorithms {effectively reduce forgetting} only with the additional assumption that \emph{learning is localised}, which we dub \emph{the locality assumption}.

\begin{assumption}[Locality assumption] 
\label{locality-assumption} Given a local approximation with $\epsilon$-region $\Omega^\epsilon_t$ for task $t$ and arbitrary $\epsilon$, the following condition holds while learning task $t+1$:
\begin{equation}
    \bm\theta \in \Omega^\epsilon_1 \cap \dots \cap \Omega^\epsilon_t
\end{equation}
\end{assumption}

Simply put, for a continual learning algorithm producing the sequence of solutions $\bm\theta_1, \dots, \bm\theta_t$, the locality assumption requires that all the solutions lie relatively close to each other. The error of a local task loss approximation is higher the farther away from the task solution. Thus, for a given error tolerance $\epsilon$ the locality assumption is broken when the distance between the task solutions is ``too high". In \cref{experiments}, we break the locality assumption by artificially increasing the distance between task solutions, and we show that local algorithms struggle in this setting.



\section{Case study: local polynomial approximations}
\label{local-poly-apprx}
A general example of local approximation of $L_t(\bm\theta)$ is the Taylor expansion around $\bm\theta_t$:
\begin{equation}
\label{taylor-p}
    \hat{L}_t(\bm\theta) = L_t(\bm\theta_t) +  {(\bm\theta - \bm\theta_t)}^\intercal \bm\nabla {L}_t(\bm\theta_t) + \frac{1}{2}\,{(\bm\theta - \bm\theta_t)}^\intercal {\dfrac{\partial^2 \mathcal{L}_t(\bm{\theta}_t)}{\partial \bm{\theta} {\partial \bm{\theta}}^\intercal}}(\bm\theta - \bm\theta_t) + \dots
\end{equation}
The approximation error for a $p$-order approximation is upper bounded by $O(\|\bm\theta - \bm\theta_t\|^p)$, and thus the $\epsilon$-region is simply a $p$-norm ball around $\bm\theta_t$ with radius proportional bounded by $\epsilon$: $\Omega^\epsilon_t = \{\bm\theta: C \cdot \|\bm\theta - \bm\theta_t\|^p < \epsilon\}$. A convenient property of this approximation is that it allows us to express forgetting in terms of the loss derivatives in $\bm\theta_t$. Indeed forgetting is simply ${E}_t(\bm\theta) = \hat{L}_t(\bm\theta) - L_t(\bm\theta_t)$. 

\subsection{Quadratic local approximations}
A wealth of studies have shown that for over-parameterized networks 
the loss tends to be very well-behaved and almost convex in a reasonable neighborhood around the minima \citep{saxe2013exact, choromanska2015loss, jacot2020neural}. Therefore in such cases a quadratic approximation of $L_t(\bm\theta)$ might be accurate.\\
If we write the Hessian matrix ${\dfrac{\partial^2 \mathcal{L}_t(\bm{\theta}_t)}{\partial \bm{\theta} {\partial \bm{\theta}}^\intercal}}$ as $\mathbf{H}_t^\star$ and denote the change in parameters due to learning a new task $\bm\theta_t - \bm\theta_{t-1}$ as $\bm\Delta_t$, following \cref{taylor-p} we can write the forgetting on task $o$ after learning $t$ as:  
\begin{equation}
\label{forgetting-quadratic}
    E_o(\bm\theta_t) = {(\bm\theta_t - \bm\theta_o)}^\intercal \bm\nabla {L}_o(\bm\theta_o) + \frac{1}{2}\,{(\bm\theta_t - \bm\theta_o)}^\intercal \mathbf{H}_o^\star (\bm\theta_t - \bm\theta_o)
\end{equation}
In order to highlight the contribution of the latest parameter update to the average forgetting $E(t) = \dfrac{1}{t-1} \sum_{o=1}^{t-1} E_o(\bm\theta_t)$, we can formulate it in a recursive fashion:
\begin{equation}
\begin{aligned}
    \label{eqt:forgetting-rec}
    {E}(t)\, = \,
    &\frac{t-1}{t}\cdot{E}(t-1) \, + \, \underbrace{\frac{1}{t}\cdot \bm\Delta_t^\intercal \left(\sum_{o=1}^{t-1}\bm\nabla{L}_o(\bm\theta_o)\right) +  \frac{1}{2t}\cdot\bm\Delta_t^\intercal \left(\sum_{o=1}^{t-1}\mathbf{H}_o^\star \right)\bm\Delta_t +\frac{1}{t} \bm{v}^\intercal\bm{\Delta}_t}_{\text{Additional forgetting due to learning task $t$, i.e., }\bm\Delta_t}
\end{aligned}
\end{equation}
where $\bm{v}^\intercal$ denotes the vector $\sum_{o=1}^{t-2}{(\bm\theta_{t-1} - \bm\theta_o)}^\intercal \mathbf{H}_{o}^\star$. 

\textbf{Example. }Let us consider the case of $t=2$ as an example, for which $\bm{v} = \bm{0}$. The forgetting of the first task after learning the second is simply: $E(2) = \bm\Delta_2^\intercal \bm\nabla{L}_1(\bm\theta_1) +  \bm\Delta_2^\intercal \mathbf{H}_1^\star \bm\Delta_2$. Putting everything together, the objective when learning the second task is: 
\begin{equation}
\label{two-tasks-obj}
    \min_{\bm\Delta_2 \in \Theta} \{L_2(\bm\theta_1 + \bm\Delta_2) + \bm\Delta_2^\intercal \bm\nabla{L}_1(\bm\theta_1) +  \bm\Delta_2^\intercal \mathbf{H}_1^\star \bm\Delta_2\} 
\end{equation}
Notice that minimizing this objective with respect to $\bm\Delta_2$ is equivalent to minimizing the multi-task loss $L_2 + L_1$. If we now assume that $\bm\theta_1$ is a \emph{local minima} of $L_1$, we get $\bm\nabla{L}_1(\bm\theta_1) = 0$ and positive semi-definite Hessian $\mathbf{H}_1^\star \succcurlyeq 0$, from which it follows that $E(2) \geq 0$. In this case the objective \cref{two-tasks-obj} can be rewritten as: 
\begin{equation}
\begin{aligned}
\label{two-tasks-obj-2}
    &\min_{\bm\Delta_2 \in \Theta} L_2(\bm\theta_1 + \bm\Delta_2) 
     \,\,\,\,\, s.t. \,\,\, \bm\Delta_2^\intercal \mathbf{H}_1^\star \bm\Delta_2 = 0
\end{aligned}
\end{equation}
Repeating the same procedure for every following task ($t>2$) we can write the optimal learning objective for any new task under a quadratic local approximation of the loss. We state our result in \cref{prop:1}.

\begin{theorem}[Optimal quadratic local continual learning]
\label{prop:1}
For any continual learning algorithm producing a sequence of parameters $\bm\theta_1, \dots, \bm\theta_t$ such that $\bm\theta_i$ is a local minima of $L_i$ and $\sup_{\bm\theta_i, \bm\theta_k} \|\bm\theta_i - \bm\theta_k\|^3 < \epsilon$ the following relationship holds: 
\begin{equation}
    {E}(1), \dots, {E}(t-1)=0 \implies {E}(t) = \frac{1}{2}\bm\Delta_t^\intercal\bigg(\frac{1}{t} \cdot \sum_{o=1}^{t-1}\mathbf{H}_o^\star\bigg)\bm\Delta_t \ge 0
\end{equation} 
Moreover, if ${E}(1), \dots, {E}(t-1)=0$ the optimal learning objective for task $t$ is: 
\begin{equation}
\begin{aligned}
\label{multi-tasks-obj-2}
    &\min_{\bm\Delta_t \in \Theta} L_t(\bm\theta_{t-1} + \bm\Delta_t) 
     \,\,\,\,\, s.t. \,\,\, \bm\Delta_t^\intercal \bigg(\frac{1}{t} \cdot \sum_{o=1}^{t-1}\mathbf{H}_o^\star\bigg) \bm\Delta_t = 0
\end{aligned}
\end{equation}
\end{theorem}

There is an intuitive explanation of \cref{prop:1}: when a quadratic approximation to the loss is accurate enough and each task solution is a local minimum, the parameter updates must be taken along directions where the multi-task loss landscape is (absolutely) flat in order to prevent forgetting. More formally, the constraint in \cref{multi-tasks-obj-2} is forcing the update $\bm\Delta_t$ to lie in the \emph{null-space} of the \emph{average Hessian matrix} $\overline{\mathbf{H}}^\star_{< t}:=\frac{1}{t}\,\sum_{o=1}^{t-1}\mathbf{H}_o^\star$.

At this point, it is natural to ask whether there exists a solution to the objective given by \cref{prop:1}. Previous studies \citep{sagun2016eigenvalues,sagun2017empirical} have observed that the loss landscape of deep neural networks is mostly degenerate around local optima, implying that most of the eigenvalues of the loss Hessian lie near zero. In other words, the rank of $\mathbf{H}_o^\star$ is rather small, and as shown theoretically in the case of deep linear networks~\citep{singh2021analytic}, this is of the order square-root the number of parameters. In the Appendix (\cref{fig:HRRM}) we plot the effective rank of the multi-task loss Hessian matrix as a function of $t$ 
and we find in practice that the rank of $\overline{\mathbf{H}}^\star_{< t}$ grows sub-linearly in $t$.  

Nevertheless, there may be settings for which the condition in \cref{multi-tasks-obj-2} is unsatisfiable (e.g. the multi-task loss Hessian matrix is full rank) or leads to poor task solutions (thereby violating the assumptions of \cref{prop:1}). \cref{eqt:forgetting-rec} describes forgetting in a more general case with no other assumptions than locality. Generally, for quadratic local approximations of the loss, the optimal learning objective for a given task is: 
\begin{equation}
\label{general-opt-local}
    \min_{\bm\Delta_t \in \Theta} \, \{ \,\,L_t(\bm\theta_{t-1} + \bm\Delta_t) + \,\bm\Delta_t^\intercal (\overline{\bm\nabla{L}}_{<t}+\frac{1}{t}\,\bm{v}) +  \frac{1}{2}\,\bm\Delta_t^\intercal \overline{\mathbf{H}}^\star_{< t}\bm\Delta_t \,\,\} 
\end{equation}
where we have introduced the abbreviation
$\overline{\bm\nabla{L}}_{<t} := \frac{1}{t}\,\sum_{o=1}^{t-1}\bm\nabla{L}_o(\bm\theta_o)$. Notice that \cref{general-opt-local} also favours parameter updates which lie in the null space of the multi-task loss Hessian matrix. However while \cref{multi-tasks-obj-2} employs a hard constraint effectively reducing the space of solutions, \cref{general-opt-local} employs a soft constraint, potentially trading-off forgetting with better performance on the task. In \cref{literature-examples}, we review examples of existing algorithms that implement the hard and soft constraints. 

The role of the multi-task loss Hessian matrix in the local learning objectives (\cref{multi-tasks-obj-2,general-opt-local}) explains the observation in the literature that \emph{"flatter local minima"}~\citep{keskar2016large}  result in reduced forgetting \citep{deng2021flattening,mirzadeh2020linear}. Under a local quadratic approximation, the flatter the previous task minima, the larger the space of solutions to the current task where forgetting is $0$ or close to $0$. Therefore \emph{continual learning algorithms favouring flatter landscapes such as \citep{deng2021flattening} implicitly rely on a local approximation of the task loss}, and thereby they are successful strategies only when learning is localised.

\section{Local and Global algorithms in the literature }
\label{literature-examples}
We now review some existing continual learning algorithms, classifying them into local and global algorithms.  
We select a few well known exemplars, representing different families of algorithms according to popular taxonomies of the literature \citep{de2021continual}. 

\subsection{Global algorithms}

\textbf{Experience Replay} (ER) is one of the oldest \citep{robins1995catastrophic} and still one of the most effective \citep{buzzega2021rethinking} algorithms for continual learning. Although several variants have been proposed   \citep{rebuffi2017icarl,lopez2017gradient,shin2017continual,van2020brain} for now we consider its simplest form. For each task, a random subset of the dataset $S_t \subset D_t$ is stored in an external buffer
to approximate the task loss $L_t(\bm\theta)$ as follows:  
\begin{equation}
\label{replay-approx}
    \hat{L}_t(\bm\theta) = \frac{1}{|S_t|} \sum_{x,y\in S_t} l_t(x,y,\bm\theta)
\end{equation}
The objective of each learning step is that of \cref{approxmtloss}. As long as the buffer sampling strategy does not depend on the network parameters after learning the task, the approximation used by experience replay is global. The approximation error is a function of the buffer size and it has been observed that in most cases the algorithm is effective also when the buffer size is small \citep{buzzega2020dark, buzzega2021rethinking}. 

\textbf{Gradient Episodic Memory} (GEM) \citep{lopez2017gradient, chaudhry2018efficient} employs the same task loss approximation in a different way. Instead of directly minimizing the multi-task loss, the objective of GEM is: 
\begin{equation}
\begin{aligned}
\label{GEM-obj}
    &\min_{\bm\Delta_t \in \Theta} L_t(\bm\theta_{t-1} + \bm\Delta_t) \quad
     \,\,\,\,\, s.t. \,\,\, 
    \hat{L}_o(\bm\theta_{t-1} + \bm\Delta_t) \le \hat{L}_o(\bm\theta_{t-1}) \,\,\, \forall\,o<t
\end{aligned}
\end{equation}
In other words, the parameter update $\bm\Delta_t$ does not minimise the approximate task loss $\hat{L}_o$ but it does not increase it (thus avoiding catastrophic forgetting). In order to enforce the constraint in \cref{GEM-obj}, GEM uses a local linearization of the 
old task loss, which is updated after each optimization step.  The linearization may be inaccurate when learning with large gradient step sizes, and result in reduced performance. Nevertheless, the task loss approximation is based on the current state of the network rather than its state after learning the task, which makes this algorithm global.  

\textbf{Synaptic Intelligence} (SI) \citep{zenke2017continual} uses a quadratic approximation of the old task loss, centered around the previous task solution. For a current task $t$ the approximation of $L_o$ is:
\begin{equation}
    \hat{L}_o(\bm\theta) = (\bm\theta - \bm\theta_{t-1})^\intercal \hat{H}_o (\bm\theta - \bm\theta_{t-1}) 
\end{equation}
where $\hat{H}_o$ is a diagonal matrix updated at each gradient step, 
which roughly estimates the task Hessian matrix evaluated at $\bm\theta_{t-1}$. The objective of each learning step is that of \cref{approxmtloss}. The approximation of $L_o$ is updated after each task 
based on the current state of the network $\bm\theta_t$, therefore SI also belongs to the group of global algorithms. Similarly to GEM, SI is sensitive to large step sizes, as the quadratic approximation may be inaccurate if the distance travelled in the parameter space is large. 

\textbf{Progressive Neural networks} (PNN) \citep{rusu2016progressive} belong to the category of \emph{network expansion} methods, which dynamically allocate new parameters of the neural network to each task. Specifically, PNN subsequently adds `\emph{columns}' (parametrized by feed-forward networks) to the neural network with unilateral connections between them. Although the whole network is used to produce outputs, only the parameters of the last column are trained on the current task, while the others are \emph{frozen}. In more general terms, let $\Theta^1, \dots, \Theta^t$ denote the parameter subspace associated to each task. By design the modified derivative of the task loss $L_t$ is: 
\begin{equation}
\dfrac{\partial \tilde{L}_t(\bm\theta)}{\partial \bm\theta_i} = 
\begin{cases}
 \dfrac{\partial L_t(\bm\theta)}{\partial \bm\theta_i} &\text{if $\bm\theta_i \in \Theta^t$}\\
0 &\text{if  $\bm\theta_i \notin \Theta^t$}
\end{cases}
\end{equation}
For all parameters $\Theta^b$ with $b>t$, both the modified derivative and the true derivative of $L_t$ are $0$ (because the parameters do not enter the output computation). However for all parameters $\Theta^b$ , $b<t$, the true derivative might be non zero. As a consequence, the approximation $\tilde{L}_t$ is more inaccurate as $t$ increases. However, forgetting is fixed to zero because the parameters are split among tasks. More precisely, a change in the parameters $\bm\theta_i \in \Theta_t$ has no effect on the forgetting on task $o<t$: ${\partial E_o(\bm\theta)}/{\partial \bm\theta_i} = {\partial (L_o(\bm\theta) - L_o(\bm\theta_o))}/{\partial \bm\theta_i} = 0.$ As a consequence, it is not only impossible to forget but it is also impossible to \emph{improve} the performance on previous tasks (when this happens we say there is \emph{backward transfer}). Finally, note that the PNN approximation uses the current state of the network to compute the modified derivative, therefore the algorithm is global. 

Other algorithms effectively partitioning the parameter space into task-specific subspaces   (such as \citep{mallya2018packnet,van2018spacenet, serra2018overcoming}) are functionally equivalent to PNN. In particular, this family of approaches avoids catastrophic forgetting by confining the parameter update to a task-specific subspace, at the cost of potentially suboptimal task solutions 
, and no backward transfer. From this point of view, an interesting parallel emerges between this family of algorithms and the general algorithm described by \cref{prop:1}
, with the critical difference that the algorithms of \cref{prop:1} are local, while methods like PNN are global.

\subsection{Local algorithms}

\textbf{Second-order regularization.}
\citet{yin2020optimization} have recently demonstrated that a large group of continual learning algorithms which use quadratic {regularizers} to prevent forgetting effectively employ a second order local approximation of the tasks loss function (discussed in \cref{local-poly-apprx}), which shows that these algorithms are local.  
More specifically, Elastic Weight Consolidation (EWC) \citep{kirkpatrick2017overcoming} and Structured Laplace Approximation \citet{ritter2018online} 
minimize the optimal objective of \cref{general-opt-local}, and differ in their assumptions on the hessian matrix.  

\textbf{iCarl} \citep{rebuffi2017icarl} uses experience replay to avoid catastrophic forgetting. There are several elements contributing to the final form of the algorithm, such as the use of network-generated targets instead of labels to approximate the task loss, and the non-parametric classifier based on the network features. Crucially, the samples in the task replay buffer are selected after the task has been learned in order to best approximate the true feature class mean. The selection procedure, based on herding, inevitably depends on the value of the network parameters after learning the task, as the network features are determined by the parameters. Substantial transformations of the feature map 
will impact the task-loss approximation accuracy. Therefore, iCarl belongs to the set of local algorithms due to this prioritized buffer selection procedure. In \cref{experiments} we experimentally verify our claim by changing the algorithm's sampling strategy.  

\textbf{Orthogonal gradient descent} (OGD) \citep{farajtabar2020orthogonal} avoids catastrophic forgetting by enforcing orthogonality between the parameter update and the previous tasks output-gradients. In order to do so efficiently, a set of the task output-gradient vectors is stored in a buffer at the end of the task. By doing so, the algorithm uses a local approximation of the task loss function. In \cref{prop:ogd} we show that OGD implements the optimal continual learning objective under local quadratic approximations of the loss (\cref{prop:1}). 
\begin{theorem}
\label{prop:ogd}
Let $\bm\Delta_1, \dots, \bm\Delta_t$ be the sequence of updates produced by Orthogonal Gradient Descent on a sequence of $t$ tasks and let $\bm\theta_1, \dots, \bm\theta_t$ be the corresponding parameters. If $\sup_{\bm\theta_i, \bm\theta_k} \|\bm\theta_i - \bm\theta_k\|^3 < \epsilon$, then $\bm\Delta_t$ satisfies the constraint in \cref{multi-tasks-obj-2} and $E_t = 0$ for all learning steps $t$. 
\end{theorem}
{This result establishes a direct link between OGD and second-order regularization methods such as EWC: in settings where a quadratic approximation of the task loss is accurate, both methods minimize the optimal learning objective, with the difference that OGD relies on a hard constrain (\cref{multi-tasks-obj-2}) and regularization methods use soft constraints (\cref{general-opt-local}). As a consequence, OGD implements a more conservative approach, potentially sacrificing performance on new tasks in order to maintain performance on old tasks, whereas EWC and the like strike a balance between new and old tasks performance which is determined by the regularization strength hyperparameter.}

\section{Experiments}
\label{experiments}

\paragraph{Experimental setup.}
With our experiments 
we want to evaluate the practical side of the theoretical distinction between local and global algorithms.
We take several classic algorithms in the literature which are representative of the different families of algorithms, namely: Expericence replay (ER), Averaged GEM (A-GEM, a computationally cheap variant of GEM), Elastic Weight Consolidation (EWC), Synaptic Intelligence (SI), iCarl and Orthogonal Gradient Descent (OGD), which were discussed in \cref{literature-examples}. We do not include Progressive Neural networks (PNN) in the experiments because we are interested in forgetting, which by design is always $0$ for PNN. 

We employ the benchmarking codebase developed by \citet{boschini2022class, buzzega2020dark}\footnote{The codebase is publicly available at https://github.com/aimagelab/mammoth}. In particular, we carry out our experiments on three popular continual learning challenges: \textbf{split CIFAR-10} \citep{cifar100}, \textbf{split Tiny-ImageNet} \citep{wu2017tiny} and \textbf{Rotated-MNIST} \citep{lecun1998gradient, lopez2017gradient}. In line with the \emph{task-} and \emph{class-incremental} settings, the first two challenges consist in splitting the original dataset into $5$ and $10$ tasks, each introducing $2$ and $20$ new classes, respectively. The Rotated-MNIST challenge belongs instead to the \emph{domain-incremental} setting, and it consists of $20$ subsequent classification tasks on the same $10$ MNIST classes, where the inputs to each task are all rotated by an angle in the interval $[0, \pi)$. 
For a review of the difference between task-incremental, class-incremental and domain-incremental settings we refer the reader to \citep{hsu2018re,van2019three}. 
Since iCarl cannot be applied to the domain-IL setting, we do not include it in the Rotated-MNIST experiments. 

For each continual learning challenge, our choice of network architectures conforms to the standard practice in the literature. Specifically, for the MNIST-based challenge we employ a fully-connected network with two hidden layers, each one comprising of $100$ ReLU units. For the challenges based on CIFAR-10 and Tiny ImageNet, we employ a ResNet18 \citep{he2016deep}. We adopt the standard hyper-parameter configuration and network training setup in \citep{buzzega2020dark}, which has been selected by grid-search. We refer the reader to \cref{experimental-details} in the Appendix for a detailed characterization of the experiment configurations and hyperparameters. 

\subsection{Local VS Global}
\label{main-experiments}

Our main experiment consists in comparing local and global algorithms in settings where the locality assumption (\cref{locality-assumption}) holds and settings where it doesn't. Recall that for a local approximation of the task loss $L_t$, the $\epsilon$-region is a neighborhood of the task solution $\bm\theta_t$. For example, for a polynomial approximation of degree $p$, the $\epsilon$-region is a $p$-norm ball around $\bm\theta_t$ of radius $\epsilon$. 
Roughly, the higher the distance travelled between tasks in the parameter space, the higher the loss approximation error for local approximations. And we say that the locality assumption is broken when this error is higher than a given tolerance of $\epsilon$.

We break the locality assumption by artificially increasing the Euclidean distance between task solutions, i.e. $\|\bm\Delta_t\|_2$. There are a number of factors which determine $\|\bm\Delta_t\|_2$, including the learning rate, the number of epochs, the batch size and the dataset size. We choose to adjust the learning rate while keeping all the other factors constant. More explicitly, we repeat all our experiments for different learning rates, which are approximately equidistant on a logarithmic scale. By varying the learning rate in a wide range, we smoothly interpolate between a \emph{local learning} setting and a \emph{non-local learning} setting.

\begin{figure}
    \centering
    \includegraphics[width=\linewidth]{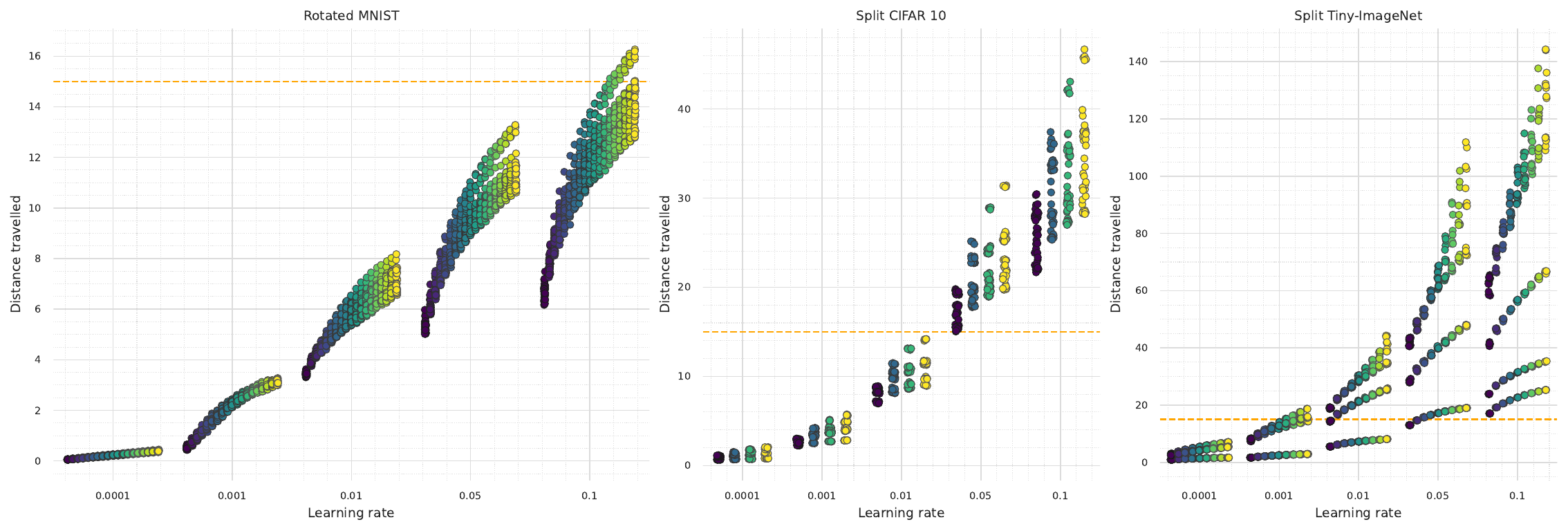}
    \caption{Distance travelled in the parameter space as a function of the optimizer learning rate and the task. We use a color coding of the tasks (a brighter color corresponding to a later task) 
    . For each task, we measure the Euclidean distance between $\bm\theta_t$ and the initialization $\bm\theta_0$.  We plot results over all algorithms and random seeds (for a total of $5$). Finally, the yellow dashed line 
    is provided as a reference of the relative scale of the $y$-axes across datasets.}
    \label{fig:updatenorm}
    \vspace{-0.2in}
\end{figure}

We verify our methodological choice by plotting the distance travelled in the parameter space as a function of the learning rate in \cref{fig:updatenorm}. For all our configurations we see that higher learning rates indeed result in a higher distance from initialization, and, consequently higher $\|\bm\Delta_t\|_2$. 
Interestingly, for low learning rates the curves look homogeneous across algorithms, whereas they diverge as the learning rate increases, suggesting that the learning dynamics between algorithms are relatively similar in the local learning setting.  
\paragraph{Main experiment.}

\begin{table}[!ht]
\vskip -0.1in
\caption{\textbf{Main experiments results.} We mark by -- non-existent experiment configurations. The $^*$ symbol on the OGD-TinyImagenet experiments indicates the use of a lower memory size (see \cref{experimental-details} for details).} 
\label{main-experiment-table}
\begin{center}
\begin{small}
\begin{sc}
\begin{tabular}{llcccccc}
\toprule
 \multirow{2}{*}{Algorithm} & \multirow{2}{*}{lr} & \multicolumn{2}{c}{\textbf{S-CIFAR-10}} & \multicolumn{2}{c}{\textbf{ROT-MNIST}} & \multicolumn{2}{c}{\textbf{S-Tiny-ImageNet}} \\
  & &  ACC $(\uparrow)$ & FGT $(\downarrow)$ & ACC $(\uparrow)$ & FGT $(\downarrow)$& ACC $(\uparrow)$ & FGT $(\downarrow)$\\
\midrule 
 \multirow{5}{*}{\textcolor{blue}{A-GEM}} & $10^{-4}$ &  $78.56_{ \, \textcolor{gray}{\pm\,  2.59}}$ & $13.39_{ \, \textcolor{gray}{\pm\,   3.48}}$ & - & - 
 & {$24.65_{\,\textcolor{gray}{\pm\,1.29}}$} & {$\bm{38.25}_{\,\textcolor{gray}{\pm\,1.21}}$} \\
  & $10^{-3}$ &  $83.65_{ \, \textcolor{gray}{\pm\,  1.94}}$ & $13.6_{ \, \textcolor{gray}{\pm\,  2.626}}$ 
  & $74.22_{ \, \textcolor{gray}{\pm\,  5.75}}$ & $\bm{12.23}_{ \, \textcolor{gray}{\pm\,  6.01}}$ 
  & {$18.12_{\,\textcolor{gray}{\pm\,0.78}}$} & {$58.41_{\,\textcolor{gray}{\pm\,0.83}}$} \\
  & $0.005$ &  - & - 
  & $79.79_{ \, \textcolor{gray}{\pm\,  5.21}}$ & $14.07_{ \, \textcolor{gray}{\pm\,  5.47}}$ & - & - \\
  & $10^{-2}$ &  
  $\bm{86.45}_{ \, \textcolor{gray}{\pm\,  2.664}}$ 
  & $\bm{12.50}_{ \, \textcolor{gray}{\pm\,  3.54}}$ 
  & $81.21_{ \, \textcolor{gray}{\pm\,  4.84}}$ & $14.41_{ \, \textcolor{gray}{\pm\,  5.12}}$ 
  & {$24.63_{\,\textcolor{gray}{\pm\,0.73}}$} & {$58.36_{\,\textcolor{gray}{\pm\,0.36}}$}  \\
  & $0.05$ &  
  $83.27_{ \, \textcolor{gray}{\pm\,  2.46}}$ 
  & $16.96_{ \, \textcolor{gray}{\pm\,  3.364}}$ 
  & $82.01_{ \, \textcolor{gray}{\pm\,  4.97}}$ & $15.98_{ \, \textcolor{gray}{\pm\,  5.26}}$ 
  & {$26.65_{\,\textcolor{gray}{\pm\,0.09}}$} & {$57.06_{\,\textcolor{gray}{\pm\,1.02}}$} \\
  & $0.1$ 
  & $80.36_{ \, \textcolor{gray}{\pm\,  1.95}}$ 
  & $20.55_{ \, \textcolor{gray}{\pm\,  2.50}}$ 
  & $\bm{82.35}_{ \, \textcolor{gray}{\pm\,   5.46}}$ & $16.07_{ \, \textcolor{gray}{\pm\,  5.72}}$ 
  & {$\bm{29.12}_{\,\textcolor{gray}{\pm\,0.69}}$} & {$54.08_{\,\textcolor{gray}{\pm\,0.90}}$} \\
\midrule 
\multirow{5}{*}{\textcolor{blue}{ER}} & $10^{-4}$ &  $78.61_{ \, \textcolor{gray}{\pm\,  1.599}}$ & $10.42_{ \, \textcolor{gray}{\pm\,  1.94}}$ & - & - 
& $34.74_{ \, \textcolor{gray}{\pm\,  0.44}}$ & $\bm{19.75}_{ \, \textcolor{gray}{\pm\,  0.66}}$ \\
  & $10^{-3}$ 
  &  $84.97_{ \, \textcolor{gray}{\pm\,  1.46}}$ 
  & $11.08_{ \, \textcolor{gray}{\pm\,  1.68}}$ 
  & $78.88_{ \, \textcolor{gray}{\pm\,  0.68}}$ & $\bm{2.57}_{ \, \textcolor{gray}{\pm\,  0.54}}$ 
  & $30.58_{ \, \textcolor{gray}{\pm\,  0.39}}$ & $40.44_{ \, \textcolor{gray}{\pm\,  0.31}}$ \\
  & $0.005$ &  - & - 
  & $85.79_{ \, \textcolor{gray}{\pm\,  0.44}}$ & $5.69_{ \, \textcolor{gray}{\pm\,  0.41}}$ 
  & - & - \\
  & $10^{-2}$ &  
  $90.42_{ \, \textcolor{gray}{\pm\,  0.69}}$ 
  & $7.35_{ \, \textcolor{gray}{\pm\,  1.08}}$ 
  & $87.03_{ \, \textcolor{gray}{\pm\,  0.74}}$ & $6.76_{ \, \textcolor{gray}{\pm\,  0.76}}$ 
  & $42.69_{ \, \textcolor{gray}{\pm\,  0.72}}$ & $35.49_{ \, \textcolor{gray}{\pm\,  0.65}}$ \\
  & $0.05$ &  
  $ \bm{91.98}_{ \, \textcolor{gray}{\pm\,  0.93}} $ 
  & $\bm{6.20}_{ \, \textcolor{gray}{\pm\,  0.92}}$ 
  & $87.99_{ \, \textcolor{gray}{\pm\,  1.00}}$ & $8.91_{ \, \textcolor{gray}{\pm\,  1.07}}$ 
  & $\bm{50.78}_{ \, \textcolor{gray}{\pm\,  0.43}}$ & $30.33_{ \, \textcolor{gray}{\pm\,  0.64}}$ \\
  & $0.1$ 
  & $91.74_{ \, \textcolor{gray}{\pm\,  0.70}}$ 
  & $6.48_{ \, \textcolor{gray}{\pm\,  0.82}}$ 
  & $\bm{88.57}_{ \, \textcolor{gray}{\pm\,  1.04}}$ & $9.05_{ \, \textcolor{gray}{\pm\,  1.10}}$ 
  & ${49.20}_{ \, \textcolor{gray}{\pm\,  0.52}}$ & $31.76_{ \, \textcolor{gray}{\pm\,  0.61}}$ \\ 
\midrule 
\multirow{5}{*}{\textcolor{blue}{SI}} & $10^{-4}$ 
&  ${68.30}_{ \, \pm\, 2.67} $ 
& $\bm{27.19}_{ \, \textcolor{gray}{\pm\,  3.26}}$ 
& - & - & $14.62_{ \, \textcolor{gray}{\pm\,  1.21}}$ & $48.41_{ \, \textcolor{gray}{\pm\,  1.04}}$ \\
  & $10^{-3}$ 
  &  $\bm{71.03}_{ \, \textcolor{gray}{\pm\,  3.05}}$ 
  & $29.27_{ \, \textcolor{gray}{\pm\,  3.28}}$ 
  & $70.42_{ \, \textcolor{gray}{\pm\,  3.15}}$ & $16.09_{ \, \textcolor{gray}{\pm\,  3.45}}$ & $13.82_{ \, \textcolor{gray}{\pm\,  0.93}}$ & $61.08_{ \, \textcolor{gray}{\pm\,  1.19}}$ \\
  & $0.005$ &  - & - 
  & $72.19_{ \, \textcolor{gray}{\pm\,  2.54}}$ & $21.80_{ \, \textcolor{gray}{\pm\,  2.68}}$ & 
- 
& 
- \\
  & $10^{-2}$ &  
  $63.91_{ \, \textcolor{gray}{\pm\,  1.71}}$ 
  & $40.29_{ \, \textcolor{gray}{\pm\,  2.24}}$ 
  & $73.83_{ \, \textcolor{gray}{\pm\,  2.76}}$ & $21.78_{ \, \textcolor{gray}{\pm\,  2.88}}$ & $17.84_{ \, \textcolor{gray}{\pm\,  1.64}}$ & $58.53_{ \, \textcolor{gray}{\pm\,  1.10}}$ \\
  & $0.05$ &  
  $ 66.66_{ \, \textcolor{gray}{\pm\,  4.77} }$ 
  & $35.99_{ \, \textcolor{gray}{\pm\,  5.97}}$ 
  & $81.26_{ \, \textcolor{gray}{\pm\,  2.57}}$ & $15.41_{ \, \textcolor{gray}{\pm\,  2.57}}$ & $45.35_{ \, \textcolor{gray}{\pm\,  2.30}}$ & $20.89_{ \, \textcolor{gray}{\pm\,  2.85}}$ \\
  & $0.1$ 
  & $68.20_{ \, \textcolor{gray}{\pm\,  3.94}}$ 
  & $33.75_{ \, \textcolor{gray}{\pm\,  5.08}}$ 
  & $\bm{85.08}_{ \, \textcolor{gray}{\pm\,  2.22}}$ & $\bm{10.76}_{ \, \textcolor{gray}{\pm\,  2.14}}$ & $\bm{52.83}_{ \, \textcolor{gray}{\pm\,  3.96}}$ & $\bm{12.91}_{ \, \textcolor{gray}{\pm\,  4.11}}$ \\
\midrule 
\multirow{5}{*}{\textcolor{red}{EWC}} & $10^{-4}$ 
&  $61.78_{ \, \textcolor{gray}{\pm\,  3.31}}$ 
& $\bm{32.63}_{ \, \textcolor{gray}{\pm\,  4.47}}$ 
& - & - & $13.08_{ \, \textcolor{gray}{\pm\,  1.47}}$ & $\bm{39.33}_{ \, \textcolor{gray}{\pm\,  2.10}}$ \\
  & $10^{-3}$ 
  &  $61.40_{ \, \textcolor{gray}{\pm\,  3.41}}$ 
  & $39.90_{ \, \textcolor{gray}{\pm\,  4.85}}$ 
  & $70.91_{ \, \textcolor{gray}{\pm\,  2.96}}$ & $\bm{15.42}_{ \, \textcolor{gray}{\pm\,  3.23}}$ & $11.96_{ \, \textcolor{gray}{\pm\,  0.51}}$ & $58.29_{ \, \textcolor{gray}{\pm\,  0.99}}$ \\
  & $0.005$ &  - & - & 
  $72.40_{ \, \textcolor{gray}{\pm\,  2.65}}$ & $21.52_{ \, \textcolor{gray}{\pm\,  2.79}}$& - & - \\
  & $10^{-2}$ &  
  $62.29_{ \, \textcolor{gray}{\pm\,  3.09}}$ 
  & $40.73_{ \, \textcolor{gray}{\pm\,  4.56}}$ 
  & $73.18_{ \, \textcolor{gray}{\pm\,  3.53}}$ & $22.54_{ \, \textcolor{gray}{\pm\,  3.72}}$ & $13.99_{ \, \textcolor{gray}{\pm\,  0.53}}$ & $64.59_{ \, \textcolor{gray}{\pm\,   0.75}}$ \\
  & $0.05$ &  
  $ \bm{62.50}_{ \, \textcolor{gray}{\pm\,  7.20}}$ 
  & $\bm{28.69}_{ \, \textcolor{gray}{\pm\,  14.03}}$ 
  & $75.18_{ \, \textcolor{gray}{\pm\,  4.42}}$ & $22.88_{ \, \textcolor{gray}{\pm\,  4.59}}$ & $16.17_{ \, \textcolor{gray}{\pm\,  1.29}}$ & $62.88_{ \, \textcolor{gray}{\pm\,  1.14}}$ \\
  & $0.1$ 
  & $54.01_{ \, \textcolor{gray}{\pm\,  1.99}}$ 
  & $43.56_{ \, \textcolor{gray}{\pm\,  11.95}}$ 
  & $\bm{75.89}_{ \, \textcolor{gray}{\pm\,  3.76}}$ & $22.57_{ \, \textcolor{gray}{\pm\,  3.89}}$ & $\bm{17.18}_{ \, \textcolor{gray}{\pm\,  1.71}}$ & $58.76_{ \, \textcolor{gray}{\pm\,  1.77}}$ \\
\midrule 

\multirow{5}{*}{\textcolor{red}{iCarl}} & $10^{-4}$ 
&  $78.92_{ \, \textcolor{gray}{\pm\,  0.68}}$ 
& $2.49_{ \, \textcolor{gray}{\pm\,  0.64}}$ 
& - & - 
& $15.91_{ \, \textcolor{gray}{\pm\,  1.19}}$ & ${1.60}_{ \, \textcolor{gray}{\pm\,  0.73}}$ \\
  & $10^{-3}$ 
  &  $92.45_{ \, \textcolor{gray}{\pm\,  0.36}}$ 
  & $\bm{0.97}_{ \, \textcolor{gray}{\pm\,  0.30}}$ 
  & - & - & $31.05_{ \, \textcolor{gray}{\pm\,  0.70}}$ & $\bm{1.23}_{ \, \textcolor{gray}{\pm\,  0.30}}$ \\
  
  & $0.005$ &  - & - & - & - & - & - \\
  & $10^{-2}$ &  
  
  $\bm{93.46}_{ \, \textcolor{gray}{\pm\,  0.55}}$ 
  & $3.31_{ \, \textcolor{gray}{\pm\,  0.56}}$ 
  & - & - 
  & $54.13_{ \, \textcolor{gray}{\pm 0.20}}$ & $6.08_{ \, \textcolor{gray}{\pm 0.32}}$ \\
  & $0.05$ &  
  $ 91.25_{ \, \textcolor{gray}{\pm\,  0.78}} $ 
  & $5.64_{ \, \textcolor{gray}{\pm\,  1.17}}$ 
  & - & - 
  & $59.45_{ \, \textcolor{gray}{0.36}}$ & $11.71_{ \, \textcolor{gray}{0.33}}$ \\
  & $0.1$ 
  & $89.76_{ \, \textcolor{gray}{\pm\,  1.00}}$ 
  & $5.05_{ \, \textcolor{gray}{\pm\,  1.56}}$ 
  & - & - 
  & $\bm{59.93}_{ \, \textcolor{gray}{0.68}}$ & $13.76_{ \, \textcolor{gray}{0.44}}$ \\
\midrule 
\multirow{5}{*}{\textcolor{red}{OGD}} & $10^{-4}$ 
&  $\bm{65.87}_{ \, \textcolor{gray}{\pm\,  4.28}}$ 
& $\bm{27.26}_{ \, \textcolor{gray}{\pm\,  5.37}}$ 
& - & - &
{${12.04}_{ \, \textcolor{gray}{\pm\,  1.97}}^*$} & {$\bm{40.82}_{ \, \textcolor{gray}{\pm\,  0.54}}^*$} \\
  & $10^{-3}$ 
  &  $64.99_{ \, \textcolor{gray}{\pm\,  4.46}}$ 
  & $35.99_{ \, \textcolor{gray}{\pm\,  4.98}}$ 
  & $\bm{70.43}_{ \, \textcolor{gray}{\pm\,  3.22}}$ & $\bm{16.09}_{ \, \textcolor{gray}{\pm\,  3.53}}$ 
& {${12.45}_{ \, \textcolor{gray}{\pm\,  0.81}}^*$} & {${59.29}_{ \, \textcolor{gray}{\pm\,  1.23}}^*$} \\
  & $0.005$ &  - & - 
  & $70.23_{ \, \textcolor{gray}{\pm\,  3.60}}$ & $24.01_{ \, \textcolor{gray}{\pm\,   3.78}}$ & - & - \\
  & $10^{-2}$ &  
  $62.64_{ \, \textcolor{gray}{\pm\,  4.23}}$ 
  & $41.90_{ \, \textcolor{gray}{\pm\,  5.40}}$ 
  & $69.78_{ \, \textcolor{gray}{\pm\,  4.65}}$ & $26.36_{ \, \textcolor{gray}{\pm\,  4.86}}$
  & {${16.01}_{ \, \textcolor{gray}{\pm\,  0.67}}^*$} & 
{${64.84}_{ \, \textcolor{gray}{\pm\,  0.23}}^*$} \\
  & $0.05$ &  
  $ 61.02_{ \, \textcolor{gray}{\pm\,  7.35}} $ 
  & $40.87_{ \, \textcolor{gray}{\pm\,  7.91}}$ 
  & $68.66_{ \, \textcolor{gray}{\pm\,  5.71}}$ & $30.02_{ \, \textcolor{gray}{\pm\,  5.98}}$ 
& {${17.45}_{ \, \textcolor{gray}{\pm\,  0.18}}^*$} & {${64.01}_{ \, \textcolor{gray}{\pm\,  0.32}}^*$} \\
  & $0.1$ 
  & $64.66_{ \, \textcolor{gray}{\pm\,  6.45}}$ 
  & $37.70_{ \, \textcolor{gray}{\pm\,  7.57}}$ 
  & $69.06_{ \, \textcolor{gray}{\pm\,  5.28}}$ & $30.04_{ \, \textcolor{gray}{\pm\,  5.51}}$ 
  & {$\bm{19.97}_{ \, \textcolor{gray}{\pm\,  1.11}}^*$} & {${59.56}_{ \, \textcolor{gray}{\pm\,  1.56}}^*$} \\
\bottomrule \\
\end{tabular}
\end{sc}
\end{small}
\end{center}
\vskip -0.1in
\end{table}


We are now ready to look at the result of our main experiment. In \cref{main-experiment-table} we report forgetting and average accuracy as a function of the learning rate for different algorithms. 
Forgetting is measured after the last task ($T$) has been learned, in terms of the task test accuracy: $\mathcal{E}^{acc}(T) = \frac{1}{T} \sum_o \operatorname{ACC}_o(\bm\theta_o) - \operatorname{ACC}_o(\bm\theta_T)$. 

Our hypothesis is that \emph{forgetting is higher for local algorithms in the non-local learning setting}, which corresponds to higher learning rates in our experiments. Further, in (\cref{literature-examples}) we claim that algorithms such as OGD, EWC, iCarl (marked in red) are local whereas algorithms such as A-GEM, ER and SI (marked in blue) are global. From our results (\cref{main-experiment-table}) we observe that local algorithms always achieve the lowest forgetting {within the the lower end of the learning rate range considered (i.e. $lr \in \{10^{-4},10^{-3}\}$) across all configurations}. On the other hand, for global algorithms, there is no discernible correlation between the learning rate and forgetting. {Alternatively, take the difference in average forgetting between the lowest two learning rates and the highest two learning rates, for each challenge. For local algorithms, the difference is always negative, whereas for global algorithms the difference can be both positive and negative.} Overall, the experiment results confirm our expectations on the functional aspects of the algorithms, namely that forgetting is higher for local algorithms when the locality assumption is not met. {This demonstrates that the difference between local and global approximations is not merely a theoretical one, as it has practical implications, observable with carefully designed experiments.}  

\begin{wrapfigure}{l}{0.3\textwidth}
\centering
\includegraphics[width=0.8\linewidth]{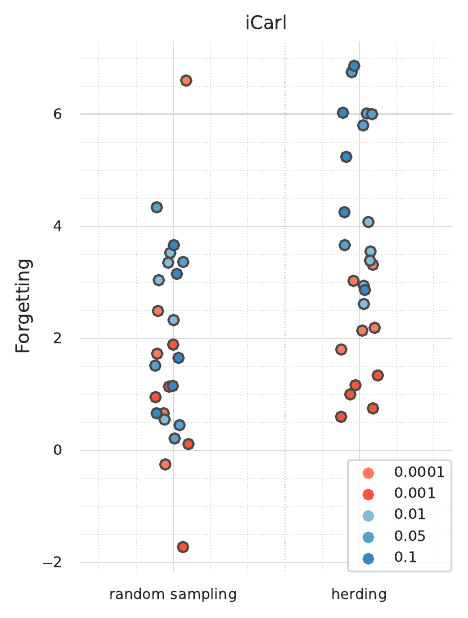}
    \caption{Comparison of random sampling (ours) and herding (standard) buffer selection strategies in iCarl. Higher learning rates, associated with non-local learning are shown in shades of blue, while learning rates associated with local learning are shown in shades of red. Each experiment is repeated over $5$ random seeds (plotted as different points).}
    \label{fig:forgetting-icarl}
\vspace{-0.2in}
\end{wrapfigure}  
In addition to these immediate conclusions, from \cref{main-experiment-table} we also observe that  average accuracy and forgetting may be at odds with each other. This contrast reflects the tension between \emph{stability and plasticity} in learning, termed the \emph{stability-plasticity dilemma} in the literature \citep{mermillod2013stability}, { which may be summarised as follows:} highly plastic but unstable models can quickly achieve high accuracy on any new task but forget as quickly the old ones; conversely, stable but rigid models avoid forgetting at the cost of lower average accuracy on all tasks. Higher learning rates (with no decay) render the network more plastic, leading to higher accuracy overall. In \cref{main-experiment-table} we see that global algorithms especially benefit from higher plasticity, as forgetting is homogeneous across learning settings. {The definition of the right balance between stability and plasticity is still an open question, and the extent to which accuracy on any past task may be sacrificed for higher average accuracy certainly depends on the application. Conceptual distinctions such as local and global algorithms can help continual learning practitioners in choosing the algorithm which is better suited to their problem.  }

\paragraph{iCarl.} 

In \cref{literature-examples} we explain that the buffer selection criteria of iCarl, based on a herding strategy, relies on the information of the task solution, which makes the loss approximation sensitive to big changes in the parameter space. In this section, we test our claims with a simple experiment. We change the buffer selection strategy of iCarl to \emph{uniformly random sampling} from the task dataset, leaving the rest of the algorithm the same. We compare the two selection strategies across multiple learning rates, following the previous experiment design.  In \cref{fig:forgetting-icarl} we plot the resulting forgetting as a function of the learning rate. 
We see that the two buffer selection strategies lead to visibly different outcomes when using high learning rates, confirming our thesis. More specifically, the herding strategy used by iCarl results in higher forgetting on average for non-local learning (shades of blue) compared to local-learning settings (shades of red). Instead using a random sampling strategy the forgetting distributions of local and non-local learning are hardly distinguishable. This experiment demonstrates the potential of the loss-approximation viewpoint for understanding continual learning algorithms, which is instrumental to address their failures.

\subsection{Robustness of local approximations}
\label{robustlocalapprox}

Finally, we take a closer look at the geometry of the parameter space around the task optima. We wish to understand how in practice the $\epsilon$-region of a task loss approximation behaves. 

To this end, we evaluate the size of the area around a local minima where a second order approximation of the loss is accurate. Recall that around a local minima the second order approximation of the loss (\cref{taylor-p}) mostly depends on the hessian matrix,  evaluated at the minima. The spectrum of this matrix is often dominated by a few, principal eigenvalues in practice. This means that, when close to the minima, moving along a principal eigenvector direction leads to the sharpest increase in the loss. However, far from the minima, where the second order approximation of the loss is no longer accurate, moving along the same direction will be no different than moving along any other direction, on average. 

\begin{figure}
    \centering
    \includegraphics[width=0.7\linewidth]{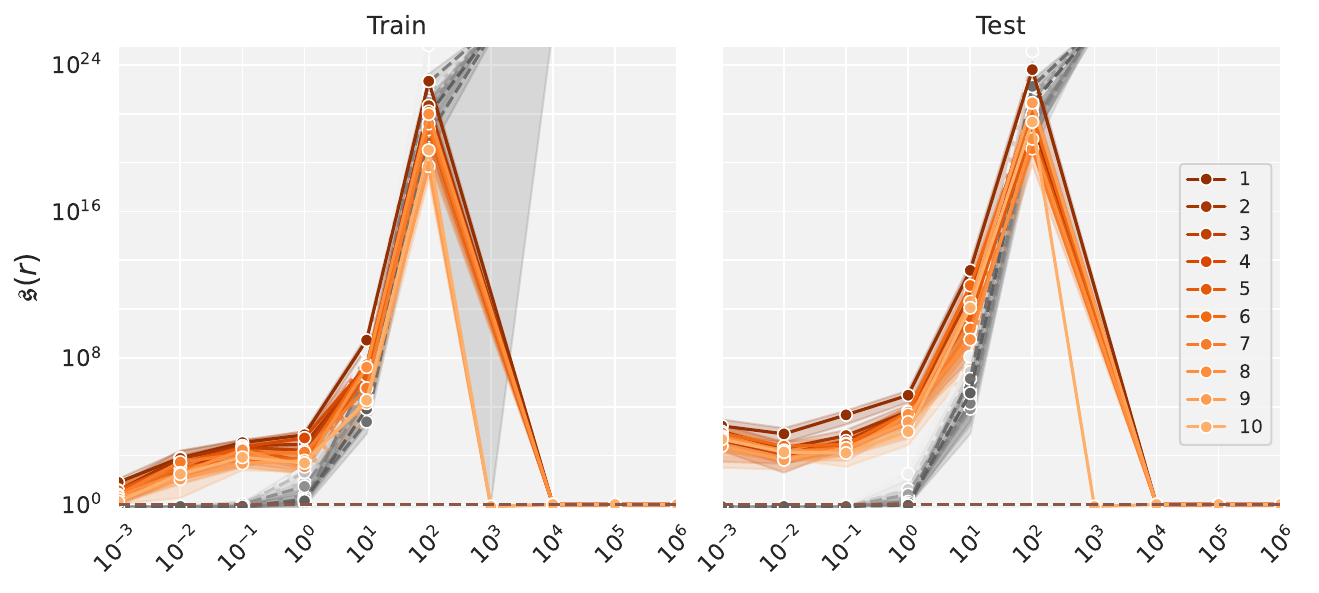}
    \caption{ 
    In \textcolor{orange}{orange}, the perturbation score $\mathfrak{s}(r)$ (\cref{score-ptb}) and in \textcolor{gray}{gray}, the task loss, evaluated on train and test data on the Split CIFAR 10 tasks. The shaded area around the curves reflects standard deviation across tasks. Different lines correspond to different perturbation directions (the first $10$ eigenvectors of the the corresponding loss). We evaluate the curves for multiple values of $r$ on a logarithmic scale in the range $[10^{-3}, 10^{6}]$.  The shape of the curve is remarkably stable across tasks. Also, notice that the score $\mathfrak{s}(r)$ on the test data is large even for $r=0$, which indicates that the test loss is not $0$ at the local optima. }
    \label{fig:perturbation}
\vskip -0.1in
\end{figure}

We observe the transition in the correctness of the second order approximation by looking at the loss function as we move away from the minima along the principal eigenvectors. We obtain the task local optima $\bm\theta_1, \dots, \bm\theta_T$ by training on each task sequentially with an SGD optimizer. 
We then evaluate the effect of perturbing the optima along a principal eigenvector through the following score: 
\begin{equation}
\label{score-ptb}
    \mathfrak{s}(r) = \mathbb{E}_{\bm{\mu}'} \left| \frac{{L}_t(\bm\theta_t + r \cdot \bm{v}_i) -  {L}_t(\bm\theta_t)} {{L}_t(\bm\theta_t + r \cdot \bm{\mu}' ) -  {L}_t(\bm\theta_t)} \right|
\end{equation}
In \cref{score-ptb} $\bm{v}_i$ represents the $i$-th principal eigenvector, and $\bm{\mu}'$ is a Gaussian random vector scaled to have unit norm. The scalar $r$ controls the distance from the optima. In \cref{fig:perturbation} we plot $\mathfrak{s}(r)$ and the loss ${L}_t(\bm\theta_t + r \cdot \bm{v}_i)$ as we vary $r$ in the range $[10^{-3}, 10^{6}]$ for all $5$ tasks of the Split CIFAR-10 challenge. We see that the score increases monotonically with the perturbation strength $r$ for low values of $r$, and it drops to $1$ after $\approx 10^3$. We can deduce that, for this specific case, the quadratic approximation is reasonably accurate in a large neighborhood of the tasks optima. Moreover, we observe that the shape of the curve is remarkably stable across tasks, which suggests that the different task optima have a similar quadratic geometry. 

\section{Related work}

We are not the first to view the continual learning problem from the loss approximation perspective. Several existing algorithms \citep{zenke2017continual, lee2016gradient, yoon2021online} have been devised in this perspective. Moreover, \citet{yin2020optimization, kong2023overcoming} have already studied regularization-based algorithms using local quadratic loss approximations.  
Differently from \citet{yin2020optimization, kong2023overcoming}, we analyze local polynomial approximations with the aim of deriving optimal continual learning objectives in local-learning setting, rather than analyzing existing algorithms.
Similarly, our results on Orthogonal Gradient Descent are also different from those of \citet{bennani2020generalisation}, who establish rigorous guarantees for OGD under the Neural Tangent Kernel~\citep{jacot2020neural} regime: our results regard instead the equivalence between OGD and the optimal objective under local quadratic approximations. 

This work is also related to existing surveys of the continual learning literature, such as \citep{de2021continual, parisi2019continual,khetarpal2022towards, qu2021recent, awasthi2019continual}, which catalogue the existing continual learning into different `families' and rank them according to different metrics. However, our classification of the literature into local and global algorithms reflects the implicit limitations of the algorithms rather than their surface-level characteristics, such as the presence of regularisation or rather the storing of input-output pairs in an external memory bank. 

Finally, our work shares the spirit of  existing theoretical studies of catastrophic forgetting and continual learning algorithms, such as \citep{mirzadeh2020understanding, mirzadeh2020linear, farquhar2019unifying, ramasesh2020anatomy, verwimp2021rehearsal}. To the best of our knowledge, we are the first to focus on the effects of the locality assumption in continual learning algorithms.

\section{Conclusion}

In summary, in this work, we view continual learning from the point of view of the multi-task loss approximation and we study the differences between local and global approximations. Based on this analysis, we provide a classification of existing algorithms into local and global algorithms, and we evaluate the practical consequences of our theoretical distinction through extensive experiments. We believe our results are of interest to both empirical and theoretical research in continual learning. In general, we find that the loss approximation viewpoint has not been sufficiently developed in the literature, and with this work, we aim to demonstrate that it offers powerful abstractions of continual learning algorithms. 

\section*{Acknowledgements}

GL is supported by a fellowship from the ETH AI Center. SPS
would like to acknowledge the financial support from Max Planck ETH Center for Learning Systems.

\newpage

\bibliography{collas2024_conference}
\bibliographystyle{collas2024_conference}

\newpage

\section*{Appendix}

\section{Proofs}
\label{AA}

\centerline{\bf Notation}
\bgroup
\def\arraystretch{1.5}
\begin{tabular}{p{1in}p{3.6in}}
$\displaystyle \mathcal{T}_1, \dots, \mathcal{T}_T$ & A sequence of tasks\\
$\mathcal{X}$ & Input space \\
$\mathcal{Y}_t$ & Task $t$ output space (may vary or be shared across tasks) \\
$\displaystyle {D}_t$ & The dataset of task $\mathcal{T}_t$: $\{(x_1,y_1),\dots,(x_{n_t},y_{n_t})\}$\\
$\Theta \subseteq \mathbb{R}^P$ & The neural network parameters \\
$\displaystyle \ra$ & A scalar \\
$\displaystyle \bm{a}$ & A vector\\
$\displaystyle \bm{A}$ & A matrix\\
$\displaystyle \mI_n$ & Identity matrix with $n$ rows and $n$ columns\\
$\bm\theta$ & A generic network parameters vector \\
$\bm{\theta}_t$ & The network parameters after training on task $t$  \\
$\bm\Delta_t := \bm\theta_t - \bm\theta_{t-1}$ & Parameter update due to the training on task $t$ \\
$\bm{\theta}_0$ & Network initialization \\
$l_t(x,y,\bm{\theta})$ & Task $t$ loss function \\
$\mathcal{L}_t(\bm{\theta})$ & Expected loss on the task $t$ distribution: $\langle l_t(x,y,\bm{\theta})\rangle_{\mathcal{P}_t}$ \\
${L}_t(\bm{\theta})$ & Empirical loss on the task $t$ dataset: $l_t(x,y,\bm{\theta})\rangle_{{D}_t}$ \\
${L}_t^\star$ & ${L}_t(\theta_t)$ \\
$\bm\nabla \mathcal{L}_t(\bm\theta)$ & Loss gradient vector: $\dfrac{\partial \mathcal{L}_t(\bm{\theta})}{\partial \bm{\theta}}$ \\
$\mathbf{H}_t(\bm{\theta})$ & Loss hessian matrix: $\dfrac{\partial^2 \mathcal{L}_t(\bm{\theta})}{\partial \bm{\theta}^2}$ \\
$\mathbf{H}_t^\star$ & $\mathbf{H}_t(\bm{\theta}_t)$ \\ 
$\overline{\bm\nabla{L}}_{<t}$ & Average gradient: $ \frac{1}{t}\,\sum_{o=1}^{t-1}\bm\nabla{L}_o(\bm\theta_o)$\\ 
$\overline{\mathbf{H}}^\star_{< t} $ & Average hessian: $\frac{1}{t}\,\sum_{o=1}^{t-1}\mathbf{H}_o^\star$\\
$\mathcal{E}_o(\bm\theta)$ & Forgetting: $\mathcal{L}_o(\bm\theta) - \mathcal{L}_o(\bm\theta_o)$ \\
$E_o(\bm\theta)$ & ${L}_o(\bm\theta) - {L}_o(\bm\theta_o)$ \\
$\mathcal{E}^{acc}_o(\bm\theta)$ & $\operatorname{ACC}_o(\bm\theta_o) - \operatorname{ACC}_o(\bm\theta)$ \\
$E(t)$ & Average forgetting: $ \frac{1}{t} \sum_{o=1}^{t} \mathcal{E}_o(\bm\theta_t)$ \\ 
$\operatorname{ACC}(t)$ & Average accuracy: $\frac{1}{t} \sum_{o=1}^{t} \operatorname{ACC}_o(\bm\theta_t)$\\
$ \mathcal{L}^{MT}_t(\bm\theta) $ & Expected multi-task loss: $ \frac{1}{t} \left( {\mathcal{L}}_1(\bm{\theta}) + \dots + {\mathcal{L}}_t(\bm{\theta})\right)$\\
$ {L}^{MT}_t(\bm\theta) $ & Empirical multi-task loss: $ \frac{1}{t} \left( {{L}}_1(\bm{\theta}) + \dots + {{L}}_t(\bm{\theta})\right)$ \\ 
$\hat{L}^{MT}_{t}(\bm\theta)$ & Approximate multi-task loss: $ \frac{1}{t} \left( {\hat{L}}_1(\bm{\theta}) + \dots + {\hat{L}}_{t-1}(\bm{\theta}) + {{L}}_t(\bm{\theta})\right)$
\end{tabular}
\egroup

\subsection{Quadratic local loss approximations (\cref{local-poly-apprx})}\label{AA_OP}

\subsubsection{\cref{eqt:forgetting-rec}}

In the following, we show the steps leading to Equation \ref{eqt:forgetting-rec}. 

We start from the formula of forgetting given by a quadratic Taylor expansion  of the task loss around $\bm\theta_t$ (Equation \ref{forgetting-quadratic}): 
\begin{align*}
    {E}_o(\bm\theta_t)
    &= {(\bm\theta_t - \bm\theta_o)}^\intercal \bm\nabla \mathcal{L}_o(\bm\theta_o) + \frac{1}{2}\,{(\bm\theta_t - \bm\theta_o)}^\intercal \mathbf{H}_o^\star(\bm\theta_t - \bm\theta_o)
\end{align*}

\begin{align*}
    {E}_o(t)
    &={(\bm\theta_t - \bm\theta_o)}^\intercal \bm\nabla \mathcal{L}_o(\bm\theta_o) + \frac{1}{2}\,\left(\bm\theta_t - \bm\theta_o\right)^\intercal\mathbf{H}_o^\star\left(\bm\theta_t - \bm\theta_o\right)  \\
    &=\left(\sum_{\tau = o+1}^t \bm\Delta_\tau\right)^\intercal \bm\nabla \mathcal{L}_o(\bm\theta_o) + \frac{1}{2}\,\left(\sum_{\tau = o+1}^t \bm\Delta_\tau\right)^\intercal\mathbf{H}_o^\star\left(\sum_{\tau = o+1}^t \bm\Delta_\tau\right)  \\
    &= \underbrace{\left(\sum_{\tau = o+1}^{t-1} \bm\Delta_\tau\right)^\intercal \bm\nabla \mathcal{L}_o(\bm\theta_o) + \frac{1}{2}\,\left(\sum_{\tau = o+1}^{t-1} \bm\Delta_\tau\right)^\intercal\mathbf{H}_o^\star\left(\sum_{\tau = o+1}^{t-1} \bm\Delta_\tau\right)}_{{E}_o(t-1)} + \bm\Delta_t^\intercal \bm\nabla \mathcal{L}_o(\bm\theta_o) + \frac{1}{2}\bm\Delta_t^\intercal\mathbf{H}_o^\star\bm\Delta_t + \\ 
    &\hspace{5pt}+\frac{1}{2}\left(\sum_{\tau = o+1}^{t-1} \bm\Delta_\tau\right)^\intercal\mathbf{H}_o^\star \bm\Delta_t 
    + \frac{1}{2}\bm\Delta_t^\intercal\mathbf{H}_o^\star\left(\sum_{\tau = o+1}^{t-1} \bm\Delta_\tau\right)  \\
    &= {E}_o(t-1) + \bm\Delta_t^\intercal \bm\nabla \mathcal{L}_o(\bm\theta_o) + \frac{1}{2}\bm\Delta_t^\intercal\mathbf{H}_o^\star\bm\Delta_t  + \left(\sum_{\tau = o+1}^{t-1} \bm\Delta_\tau\right)^\intercal\mathbf{H}_o^\star \bm\Delta_t  
\end{align*}

Based on this formulation, we characterize the average forgetting as follows: 
\begin{align*}
    \label{eqt:forgetting-rec}
    E(t)
    &= \frac{1}{t} \sum_{o=1}^t E_o(t) =  \frac{1}{t}\underbrace{{E}_t(t)}_{=0} + \frac{1}{t} \sum_{o=1}^{t-1} E_o(t) \\
    &=\frac{1}{t} \sum_{o=1}^{t-1} \left[ E_o(t-1) + \bm\Delta_t^\intercal \bm\nabla \mathcal{L}_o(\bm\theta_o) + \frac{1}{2}\bm\Delta_t^\intercal\mathbf{H}_o^\star\bm\Delta_t  + \left(\sum_{\tau = o+1}^{t-1} \bm\Delta_\tau\right)^\intercal\mathbf{H}_o^\star \bm\Delta_t  \right] \\
    &= \frac{1}{t} \sum_{o=1}^{t-1} \left[ E_o(t-1) \right] + \frac{1}{t}\,\bm\Delta_t^\intercal \bm\nabla \mathcal{L}_o(\bm\theta_o) + \frac{1}{t} \sum_{o=1}^{t-1} \left[\frac{1}{2}\bm\Delta_t^\intercal\mathbf{H}_o^\star\bm\Delta_t\right] + \frac{1}{t} \sum_{o=1}^{t-1} \left[\left(\sum_{\tau = o+1}^{t-1} \bm\Delta_\tau\right)^\intercal\mathbf{H}_o^\star \bm\Delta_t \right] \\
    &= \small{\frac{t-1}{t}}\cdot {E}(t-1)  + \frac{1}{t}\,\bm\Delta_t^\intercal \bm\nabla \mathcal{L}_o(\bm\theta_o) + \frac{1}{2t} \bm\Delta_t^\intercal\left[\sum_{o=1}^{t-1} \mathbf{H}_o^\star\right]\bm\Delta_t +  \frac{1}{t}\left[\underbrace{\sum_{o=1}^{t-1}(\bm\theta_{t-1}-\bm\theta_o)^\intercal  \mathbf{H}_o^\star}_{\bm{v}^\intercal}\right] \bm\Delta_t \\
    &= \frac{1}{t} \,\left(\, ({t-1})\cdot E(t-1) + \bm\Delta_t^\intercal \bm\nabla \mathcal{L}_o(\bm\theta_o) + \,\frac{1}{2}\bm\Delta_t^\intercal(\sum_{o=1}^{t-1}\mathbf{H}_o^\star)\bm\Delta_t +\bm{v}^\intercal\bm{\Delta}_t \,\right) 
\end{align*}


\newtheorem*{prop:theorem1}{Theorem \ref{prop:1}}
\subsubsection{Theorem \ref{prop:1} (Optimal quadratic local continual learning)}

The first part of our proof is by induction. We prove the base case for $t=2$ and $t=3$, since some terms trivially cancel out for $t=1$. 

The theorem makes the following assumptions: (1) every task solution $\bm\theta_i$ is a local minima of $L_i$ and (2) $\,\,\sup_{\bm\theta_i, \bm\theta_k} \|\bm\theta_i - \bm\theta_k\|^3 < \epsilon$. From the second assumption we deduce that the error of a local quadratic approximation of $L_i$ is bounded by an arbitrary $\epsilon$ and may therefore be safely ignored. The first assumption instead lets us state $\bm\nabla{L}_i(\bm\theta_i) = 0$ and the Hessian $\mathbf{H}_i^\star \succcurlyeq 0$ is p.s.d.

\textit{\underline{Base case 1}: $E(1)=0 \implies E(2) = \frac{1}{2}\bm\Delta_2^\intercal \bm{H}_1^\star \bm\Delta_2$}.
Notice that by definition, $E(1) = \mathcal{E}_1(1) = 0$.

Using Equation \ref{eqt:forgetting-rec} we can write $E(2)$:
\begin{align*}
    E(2)
    &= {E(1)} + \,\frac{1}{2}\bm\Delta_2^\intercal(\bm{H}_1^\star)\bm\Delta_2 \,+ (\bm\theta_{1} - \bm\theta_1)^\intercal \bm{H}_{1}^\star\bm\Delta_2\\
    &= 0 + \,\frac{1}{2}\bm\Delta_2^\intercal\bm{H}_1^\star\bm\Delta_2 \,+ {0}
\end{align*}

\textit{\underline{Base case 2}: $E(1)=0, E(2)=0 \implies E(3) = \frac{1}{2}\bm\Delta_3^\intercal (\frac{1}{3}\bm{H}_1^\star + \frac{1}{3}\bm{H}_2^\star) \bm\Delta_3$}.\\
Using the last result from case 1, we have that : 
\begin{align*}
    E(2)
    &= \,\frac{1}{2}\bm\Delta_2^\intercal\bm{H}_1^\star\bm\Delta_2 = 0
\end{align*}
The latter equation implies that $\bm\Delta_2^\intercal\bm{H}_{1}^1 = \bm{0}$.
Plugging it into the value of $E(3)$ given by Equation \ref{eqt:forgetting-rec}:
\begin{align*}
    E(3)
    &= \frac{1}{3} \bigg( 2\cdot{E(2)} + \,\frac{1}{2}\bm\Delta_3^\intercal(\bm{H}_2^\star + \bm{H}_1^\star)\bm\Delta_3 \,+ \sum_{t = 1}^{2} (\bm\theta_{2} - \bm\theta_t)^\intercal \bm{H}_{t}^\star\bm\Delta_3  \bigg)\\
    &= 0 + \,\frac{1}{2}\bm\Delta_3^\intercal(\frac{1}{3}\bm{H}_2^\star + \frac{1}{3}\bm{H}_1^\star)\bm\Delta_3 \,+ \frac{1}{3}\underbrace{\bm\Delta_2^\intercal\bm{H}_{1}^1}_{=\bm{0}}\bm\Delta_3\,+ \frac{1}{3}\underbrace{(\bm\theta_{2} - \bm\theta_2)^\intercal}_{=\bm{0}} \bm{H}_{t}^\star\bm\Delta_3
\end{align*}

\textit{\underline{Induction step}: if $E(\tau)=0 \,\, \forall \tau < t$  then $E(t)\geq 0$}.\\
We start by writing out $E(t)$. 
\begin{align*}
    E(t)
    &= \frac{1}{t} \bigg((t-1)\cdot E(t-1) + \,\frac{1}{2}\bm\Delta_t^\intercal\big(\sum_{\tau=1}^{t-1}\bm{H}_\tau^\star\big)\bm\Delta_t + \sum_{\tau = 1}^{t-1} (\bm\theta_{t-1} - \bm\theta_\tau)^\intercal \bm{H}_{\tau}^\star\,\bm\Delta_t  \bigg)\\
    & = 0 + \,\frac{1}{2} \bm\Delta_t^\intercal\big(\sum_{\tau=1}^{t-1}\bm{H}_\tau^\star\big)\bm\Delta_t + \sum_{\tau = 1}^{t-1} (\bm\Delta_{\tau+1} + \dots + \bm\Delta_{t-1})^\intercal \bm{H}_{\tau}^\star\,\bm\Delta_t
\end{align*}

The induction step is accomplished if $\sum_{\tau = 1}^{t-1} (\bm\Delta_{\tau+1} + \dots + \bm\Delta_{t-1})^\intercal \bm{H}_{\tau}^\star = \sum_{\tau = 1}^{t-2} (\bm\Delta_{\tau+1} + \dots + \bm\Delta_{t-1})^\intercal \bm{H}_{\tau}^\star = \bm0$. Consider now the case in which $\sum_{\tau = 1}^{t-1} (\bm\Delta_{\tau+1} + \dots + \bm\Delta_{t-1})^\intercal \bm{H}_{\tau}^\star \neq \bm0$. It follows that: 
\begin{align*}
    &\sum_{\tau = 1}^{t-2} (\bm\Delta_{\tau+1} + \dots + \bm\Delta_{t-1})^\intercal \bm{H}_{\tau}^\star \neq \bm0\\
    & \sum_{\tau = 1}^{t-2}\bm\Delta_{t-1}^\intercal \bm{H}_{\tau}^\star + \sum_{\tau = 1}^{t-2} (\bm\Delta_{\tau+1} + \dots + \bm\Delta_{t-2})^\intercal\bm{H}_{\tau}^\star \neq \bm0\\
\end{align*}
Multilying by $\bm\Delta_{t-1}$ on the right we get: 
\begin{align*}
    &\sum_{\tau = 1}^{t-2}\bm\Delta_{t-1}^\intercal \bm{H}_{\tau}^\tau \bm\Delta_{t-1} + \sum_{\tau = 1}^{t-3} (\bm\Delta_{\tau+1} + \dots + \bm\Delta_{t-2})^\intercal\bm{H}_{\tau}^\tau\bm\Delta_{t-1}  \neq \bm0^\intercal\bm\Delta_{t-1} 
\end{align*}
We compare this result with the value of $\mathcal{E}(t-1)$ given by Equation \ref{eqt:forgetting-rec}:
\begin{align*}
   E(t-1) =&\frac{1}{t-1} \,\bigg(\, ({t-2})\cdot \mathcal{E}(t-2) + \,\frac{1}{2}\bm\Delta_{t-1}^\intercal(\sum_{o=1}^{t-2}\mathbf{H}_o^\star)\bm\Delta_{t-1} + \sum_{\tau = 1}^{t-2} (\bm\Delta_{\tau+1} + \dots + \bm\Delta_{t-1})^\intercal \bm{H}_{\tau}^\star\,\bm\Delta_{t-1} \,\bigg)\\
    &= \frac{1}{t-1} \bigg( 0 + \frac{1}{2}\bm\Delta_{t-1}^\intercal(\sum_{\tau=1}^{t-2}\bm{H}_\tau^\star)\bm\Delta_{t-1} + \sum_{\tau = 1}^{t-2} (\bm\Delta_{\tau+1} + \dots + \bm\Delta_{t-1})^\intercal \bm{H}_{\tau}^\star\,\bm\Delta_{t-1} \big) \neq 0
\end{align*}
We arrived at a contradiction, which proves the induction step and concludes that if ${E}(1), \dots, {E}(t-1)=0$ then 
\begin{equation}
\begin{aligned}
\label{forgetting-lowerbound}
E(t) = \frac{1}{2} \bm\Delta_t^\intercal \bigg(\frac{1}{t} \cdot \sum_{o=1}^{t-1}\mathbf{H}_o^\star\bigg) \bm\Delta_t \ge 0
\end{aligned}
\end{equation}

The second part of the proof follows from the simple observation that minimizing the multi-task loss $L^{MT}_t$ with respect to $\bm\Delta_t$ is equivalent to minimizing $L_t + E(t)$ with respect to $\bm\Delta_t$. Morover, notice that, by definition, $E(1)=0$ for any continual learning algorithm, from which it follows by \cref{forgetting-lowerbound} that $E(2) \ge 0$. Thus, if there exist a parameter update such that $L_2 + E(2)$ is minimized, then $E(2)=0$ and therefore the objective can be written as $\min_{\bm\Delta_2} L_2\,\, s.t. \,\,\,E(2)=0$.
Recursively applying this argument it holds:
\begin{align*}
    &E(2) = \bm\Delta_2^\intercal(\sum_{o=1}^{1} \mathbf{H}_{o}^\star) \bm\Delta_2 = 0\\
    & \dots \\ 
    &E(t-1)= \bm\Delta_{t-1}^\intercal(\sum_{o=1}^{T-1} \mathbf{H}_{o}^\star) \bm\Delta_{t-1} = 0
\end{align*}
Then the optimal objective for the $t$ task can be written as \cref{multi-tasks-obj-2}.

\subsection{Local and Global algorithms in the literature (\cref{literature-examples})}


\subsubsection{\cref{prop:ogd} (OGD implements the local quadratic optimal objective)}
We start by recalling the OGD algorithm. Let $D_o = \{(\bm{x}_{1}, {y_{1}}),\cdots,(\bm{x}_{n_o}, {y_{n_o}}))\}$ be the dataset associated with task $\mathcal{T}_o$ and $\bm{f}_{\bm{\theta}}(\bm{x}) \in \mathbb{R}^K$ be the network output corresponding to the input $\bm{x}$. The gradient of the network output with respect to the network parameters is the $P \times K$ matrix:
\begin{equation*}
    \nabla_{\bm{\theta}} \bm{f}_{\bm{\theta}}(\bm{x}) = \left[\nabla_{\bm{\theta}} \bm{f}^1_{\bm{\theta}}(\bm{x}),\,\cdots\,, \nabla_{\bm{\theta}} \bm{f}^K_{\bm{\theta}}(\bm{x}) \right]
\end{equation*}
The standard version of OGD imposes the following constraint on any parameter update $\bm{u}$: 
\begin{equation}
\label{ogd-constraint}
    \langle \bm{u},\nabla_{\bm{\theta}} \bm{f}^k_{\bm{\theta}_o}(\bm{x}_{i}) \rangle = 0 
\end{equation}
for all $k\in[1,K]$, $\bm{x}_{i} \in D_o$ and $o\le t$, 
$t$ being the number of tasks solved so far. If SGD is used, for example, the update $\bm{u}$ is the gradient of the new task loss for a data batch. 
Notice that the old task gradients are evaluated at the minima $\bm{\theta}_o$.

In the first part of the proof we want to show that the OGD constraint (\cref{ogd-constraint}) is equivalent to the constraint in \cref{multi-tasks-obj-2}, namely $\bm\Delta_t^\intercal \bigg(\frac{1}{t} \cdot \sum_{o=1}^{t-1}\mathbf{H}_o^\star\bigg) \bm\Delta_t  = 0$. 

In \cref{prop:ogd} we make the assumption that the task solutions $\bm\theta_1, \dots, \bm\theta_t$ are local minima of the respective losses, from which it follows that $\bm\nabla{L}_t(\bm\theta_t) = 0$ and the Hessian $\mathbf{H}_t^\star \succcurlyeq 0$.

Recall the definition of the average loss:
\begin{equation*}
    E{L}_t(\bm{\theta}) = \frac{1}{n_t} \sum_{i=0}^{n_t} l_t(\bm{x}_{i,t},y_{i,t},\bm{\theta})
\end{equation*}
The Hessian matrix of the loss $\bm{H}_t(\bm\theta)$ can be decomposed as a sum of two other matrices \citep{schraudolph2002fast}: the \emph{outer-product} Hessian and the \emph{functional} Hessian.
\begin{align}
\label{eq:Hessian-decomp}
    \bm{H}_t(\bm\theta) = 
    \frac{1}{{n_t}} \sum_{i=1}^{n_t} \nabla_{\bm\theta} \bm{f}_{\bm{\theta}}(\bm{x}_i)\left[\nabla_{\bm{f}}^2 {\ell}_i\right] \nabla_{\bm{\theta}} \bm{f}_{\bm{\theta}}\left(\bm{x}_i\right)^{\intercal}  
    + 
    \frac{1}{{n_t}} \sum_{i=1}^{n_t} \sum_{k=1}^K\left[\nabla_{\bm{f}} {\ell}_i\right]_k \nabla_{\bm{\theta}}^2 \bm{f}_{\bm{\theta}}^k\left(\bm{x}_i\right)\,,
\end{align}
where $\ell_i = l(\bm{x_i}, y_i)$ and $\nabla_{\bm{f}}^2 {\ell}_i$ is the $K\times K$ matrix of second order derivatives of the loss ${\ell}_i$ with respect to the network output $\bm{f}_{\bm\theta}(\bm{x_i})$.
At the local minimum $\bm\theta_t$ the contribution of the functional Hessian is negligible \citep{singh2021analytic}. We rewrite the first goal of the proof using these two facts:
\begin{equation*}
    \bm\Delta_t^\intercal \mathbf{H}_{o}^\star \bm\Delta_t = 0
\end{equation*}
\begin{equation}
\label{eqt:opt-cond-revisited}
    \bm\Delta_t^\intercal\left(
    \frac{1}{{n_o}} \sum_{i=1}^{n_o} \nabla_{\bm\theta} \bm{f}_{\bm{\theta}}(\bm{x}_i)\left[\nabla_{\bm{f}}^2 {\ell}_i\right] \nabla_{\bm{\theta}} \bm{f}_{\bm{\theta}}\left(\bm{x}_i\right)^{\intercal}  \right) \bm\Delta_t = 0
\end{equation}
\cite{farajtabar2020orthogonal} apply the OGD constraint (\cref{ogd-constraint}) to the batch gradient vector $\bm{g}_B=\nabla\mathcal{L}_t^B$. Following this choice $\bm\Delta_t = \sum_{s=1}^{S_t} - \eta \bm{g}_s$, where $\eta$ is the learning rate. Clearly, if, for all $s$, $\bm{g}_s$ satisfies \cref{ogd-constraint}, then $\bm\Delta_t$ satisfies it. Hereafter, we ignore the specific form of $\bm\Delta_t$, proving the result for a broader class of algorithms for which $\bm\Delta_t$ satisfies the OGD constraint. Continuing from Equation \ref{eqt:opt-cond-revisited}:
\begin{align*}
    &
    \frac{1}{{n_o}} \left(
     \sum_{i=1}^{n_o} \nabla_{\bm\theta} \bm\Delta_t^\intercal \bm{f}_{\bm{\theta}}(\bm{x}_i)\left[\nabla_{\bm{f}}^2 {\ell}_i\right] \nabla_{\bm{\theta}} \bm{f}_{\bm{\theta}}\left(\bm{x}_i\right)^{\intercal} \bm\Delta_t  \right) \\
    &
    \frac{1}{{n_o}} \left(
     \sum_{i=1}^{n_o} \nabla_{\bm\theta} \left[ \bm\Delta_t^\intercal \left[\nabla_{\bm{\theta}} \bm{f}^1_{\bm{\theta}}(\bm{x}_i),\,\cdots\,, \nabla_{\bm{\theta}} \bm{f}^K_{\bm{\theta}}(\bm{x}_i) \right]\right] 
     \left[\nabla_{\bm{f}}^2 {\ell}_i\right] \nabla_{\bm{\theta}} \bm{f}_{\bm{\theta}}\left(\bm{x}_i\right)^{\intercal} \bm\Delta_t  \right)\\
    &
    \frac{1}{{n_o}} \left(
     \sum_{i=1}^{n_o} \nabla_{\bm\theta} \left[\underbrace{\bm\Delta_t^\intercal\nabla_{\bm{\theta}} \bm{f}^1_{\bm{\theta}}(\bm{x}_i)}_{=0},\,\cdots\,, \underbrace{\bm\Delta_t^\intercal\nabla_{\bm{\theta}} \bm{f}^K_{\bm{\theta}}(\bm{x}_i)}_{=0} \right]
     \left[\nabla_{\bm{f}}^2 {\ell}_i\right] \nabla_{\bm{\theta}} \bm{f}_{\bm{\theta}}\left(\bm{x}_i\right)^{\intercal} \bm\Delta_t  \right) = 0,
\end{align*}
where $\bm\Delta_t^\intercal\nabla_{\bm{\theta}} \bm{f}^k_{\bm{\theta}}(\bm{x}_i) = 0$ is the OGD constraint, which holds for any $k,i$ and $o<t$.
This concludes the first part of the proof. 

In order to conclude the proof we simply apply \cref{prop:1} to OGD, and see that, if the algorithm objective is minimized at each learning step, then necessarily it follows that $E(t)=0 \,\,\forall\, t \in [T]$.

\subsubsection{On the validity of OGD-GTL (Additional result) }
Instead of considering all the function gradients with respect to all the outputs $\{\nabla_{\bm{\theta}} \bm{f}^k_{\bm{\theta}}(\bm{x}) \,|\, k \in [1,K], \bm{x} \in D_o\}$, ~\cite{farajtabar2020orthogonal} also consider a cheaper approximation where they impose orthogonality only with respect to the index corresponding to the true ground truth label (GTL). We show this can be understood via fairly mild assumptions, if the loss function is cross-entropy.  

For a cross-entropy loss, $\bm{f}^k_{\bm\theta}(\bm{x_i})$ is the log-probability (or logit) associated with class $k$ for input $\bm{x_i}$. The probability $p(y_i = k|\bm{x_i};\bm\theta)$ is then defined as $(\bm{p_i})_j = \operatorname{softmax}(\bm{f}_{\bm\theta}(\bm{x_i}))_k$.
Recall, from the proof of Theorem \ref{prop:ogd}, that the theorem statement can be equivalently written as: 
\begin{equation*}
    \bm\Delta_t^\intercal\left(
    \frac{1}{{n_o}} \sum_{i=1}^{n_o} \nabla_{\bm\theta} \bm{f}_{\bm{\theta}}(\bm{x}_i)\left[\nabla_{\bm{f}}^2 {\ell}_i\right] \nabla_{\bm{\theta}} \bm{f}_{\bm{\theta}}\left(\bm{x}_i\right)^{\intercal}  \right) \bm\Delta_t = 0
\end{equation*}
For a cross-entropy loss the Hessian of the loss with respect to the network output is given by:
$$\nabla_{\bm{f}}^2 {\ell}_i = \operatorname{diag}(\bm{p_i})-\bm{p_i p_i}^{\intercal}$$
Without loss of generality, assume index $k=1$ corresponds to the GTL. Towards the end of the usual training, the softmax output at index $1$, i.e., $p_1\approx 1$. Let us assume that the probabilities for the remaining output coordinates is equally split between them. More precisely, let 
$$p_2, \cdots, p_K = \dfrac{1-p_1}{K-1}.$$ 
Hence, in terms of their scales, we can regard $p_1 = \mathcal{O}(1)$ while $p_2, \cdots, p_K = \mathcal{O}\left(K^{-1}\right)$. Furthermore, $(1-p_i) = \mathcal{O}(1)\quad \forall \, i \in [2\cdots K]$. 
Then, a simple computation of the inner matrix in the Hessian outer product, i.e., $\nabla_{\boldsymbol{f}}^2 {\ell}=\operatorname{diag}(\boldsymbol{p})-\boldsymbol{p p}^{\top}$, will reveal that \textit{diagonal entry corresponding to the ground truth label is $\mathcal{O}(1)$, while rest of the entries in the matrix are of the order $\mathcal{O}\left(K^{-1}\right)$ or $\mathcal{O}\left(K^{-2}\right)$} --- thereby explaining this approximation.
Also, we can see that this approximation would work well only when $K\gg1$ and does not carry over to other loss functions, e.g. Mean Squared Error (MSE). In fact, $K=2$ all the entries of this matrix are of the same magnitude $|p_1(1-p_1)|$, and for MSE simply $\nabla_{\bm{f}}^2 {\ell}_i = \bm{I}_K$.

\section{Details on the experimental setup}
\label{experimental-details}

\subsection{Experiments configuration}

We follow the same experiment configurations as \citep{buzzega2020dark}. In \cref{experiments-table} we summarise the configurations used in the main experiment (\cref{main-experiments}). 

\begin{table}[ht]
\caption{\textbf{Overview of the experiments configurations.}}
\label{experiments-table}
\vskip 0.15in
\begin{center}
\begin{small}
\begin{sc}
\begin{tabular}{lcccl}
\toprule
Data & Challenge type & Tasks & Architecture ($P$) & Algorithms \\
\midrule

{Rotated-MNIST} & Domain-IL & 20 &  {MLP ($89$K)} & A-GEM, ER, SI, EWC, OGD \\
{Split CIFAR-10} & Task-IL, Class-IL & 5 &  {ResNet18} ($11$M) & A-GEM, ER, SI, iCarl, EWC, OGD \\
{Split Tiny-ImageNet} & Task-IL, Class-IL & 20 &  {ResNet18} ($11$M) & A-GEM, ER, SI, iCarl, EWC, OGD \\
 \midrule
\bottomrule
\end{tabular}
\end{sc}
\end{small}
\end{center}
\vskip -0.1in
\end{table}

\subsection{Training setting and hyperparameters}

\paragraph{Hyperparameters}
In our experiments we use the optimal hyperparameters provided by \citet{buzzega2020dark}. They select the hyperparameters for each experiment configuration  by performing a grid-search on a validation set, the latter obtained by sampling 10\% of the training set. In \cref{hyperparameters-table} we report all the value of the optimal hyperparameters, which were used in our experiments. We mark by an asterisk the hyper-parameter which we modified in our experiment configuration compared to \citep{buzzega2020dark}. In particular, we increased the number of epochs from $1$ to $5$ in the Rotated MNIST challenge in order to train effectively with very low learning rates.

\begin{table}[ht]
\caption{\textbf{Overview of hyperparameters.}}
\label{hyperparameters-table}
\vskip 0.15in
\begin{center}
\begin{small}
\begin{sc}
\begin{tabular}{llcccc}
\toprule
Data & Algorithm & Buffer size & batch size & Epochs & Other HP\\
\midrule
\multirow{5}{*}{Rotated-MNIST} & A-GEM & 500 &  128 & 5* & - \\
 & ER & 500 &  128 & 5* &  -\\
 & SI & 500 &  128 & 5* &  $c = 1, \xi = 1$\\
 & EWC & - &  128 & 5* &  $\lambda= 0.7,\gamma= 1.0$\\
 & OGD & 500 &  128 & 5* & 'gtl' variant\\
 \midrule
\multirow{6}{*}{Split CIFAR-10} & A-GEM & 200 &  32 & 50 & - \\
 & ER & 200 &  32 & 50 &  -\\
 & SI & 200 &  32 & 50 &  $c = 0.5, \xi = 1$\\
 & EWC & - &  32 & 50 &  $\lambda= 10,\gamma= 1.0$\\
 & OGD & 200 &  32 & 50 & 'gtl' variant\\
 & iCarl & 200 &  32 & 50 & weight decay  $10e^{-6}$\\
  \midrule
\multirow{4}{*}{Split TinyImagenet} & {A-GEM} & {500} &  {32} & {100} & - \\
 & ER & 500 &  32 & 100 &  -\\
 & SI & 500 &  32 & 100 &  $c = 0.5, \xi = 1$\\
 & EWC & - &  32 & 100 &  $\lambda= 25,\gamma= 1.0$\\
 & {OGD} & {200*} &  {32} & {100} & {'gtl' variant}\\
 & iCarl & 500 &  32 & 100 & weight decay  $10e^{-6}$\\
 \midrule
\bottomrule
\end{tabular}
\end{sc}
\end{small}
\end{center}
\vskip -0.1in
\end{table}

We repeat all our experiments for $5$ different seeds, namely $11, 13, 33, 21, 55$, and $5$ different learning rates $[0.0001, 0.001, 0.01, 0.05, 0.1]$. For Rotated-MNIST $0.0001$ is too small to produce any meaningful results and we add $0.005$ to the list instead. Finally, following \citep{buzzega2020dark} we apply we apply random crops and horizontal flips to both stream and buffer examples for  CIFAR-10 and Tiny ImageNet.

The same hyperparameters used for iCarl are carried over to the variant of iCarl using random sampling which we discuss in \cref{main-experiments}.

\paragraph{Training settings.}

All networks are trained with Stochastic Gradient Descent (SGD), with a constant learning rate (as in \citep{buzzega2020dark}). Unless otherwise stated, we do not use weight decay or momentum. 

\subsection{Main experiment (\cref{main-experiments}): metrics }

In \cref{main-experiment-table} and \cref{fig:forgetting-icarl} we report average forgetting after learning the last task in the sequence. This is measured as the signed difference in test accuracy, averaged over tasks:
\begin{equation*}
    E^{acc}(T) = \frac{1}{T} \sum_{o=1}^T \operatorname{ACC}_o(\bm\theta_o) - \operatorname{ACC}_o(\bm\theta_T)
\end{equation*}
Moreover, we report the average accuracy, over all tasks after learning the last task in the sequence: 
\begin{equation*}
    ACC(T) = \frac{1}{T} \sum_{o=1}^T \operatorname{ACC}_o(\bm\theta_T)
\end{equation*}

\subsection{Robustness of local approximations (\cref{robustlocalapprox}): hessian eigenvectors computation}

In order to compute the perturbation score (\cref{robustlocalapprox}) we have to obtain the Hessian matrix first $10$ eigenvectors. We do so by using  the \emph{Lanczos method}, a cheap iterative method, and the Hessian-vector product \citep{yao2018large, xu2018accelerated}. \citet{hessian-eigenthings} has publicly provided an implementation of these methods for Pytorch neural networks, which we have adapted to our scope. We randomly sample $2000$ inputs from the dataset to compute the Hessian and/or its eigenvectors.

\section{Additional results}

\subsection{Average Hessian rank}

In \cref{local-poly-apprx} we argue that the Hessian matrix has a central role in local approximations of the task loss. In particular, the rank of the average task hessian $\overline{\mathbf{H}}^\star_{< t}  = \frac{1}{t}\,\sum_{o=1}^{t-1}\mathbf{H}_o^\star$ indicates the size of the parameter subspace where forgetting is $0$. This quantity is especially important for algorithms, such as Orthogonal Gradient Descent, which effectively constrain the parameter update to the null-space of $\overline{\mathbf{H}}^\star_{< t}$ (\cref{prop:ogd} \cref{prop:1}).  

\begin{figure}
\begin{center}
\vspace{-15pt}
\includegraphics[width=0.6\linewidth]{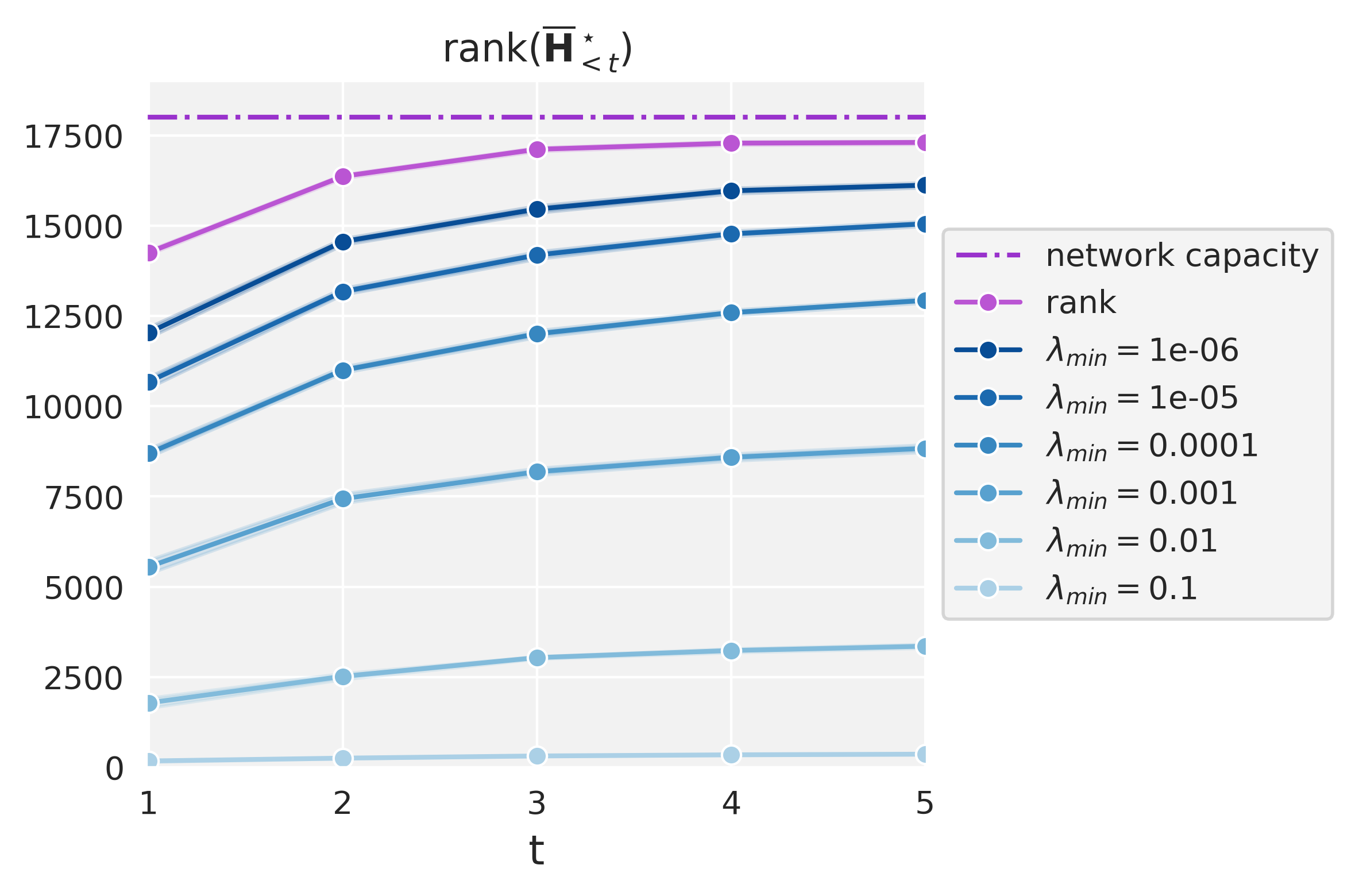}
\end{center}
    \caption{Rank and effective rank for various threshold $\lambda$ values on a tiny Rotated MNIST challenge. The values are averaged over $5$ seeds.} 
    \label{fig:HRRM}
\vspace{-10pt}
\end{figure}

In Figure \ref{fig:HRRM} we plot the evolution of $\operatorname{rank}(\overline{\mathbf{H}}^\star_{< t})$ over $t$ for a tiny version of the Rotated-MNIST challenge (with only $5$ tasks), for which we use a toy MLP network of $18$K parameters. We have to reduce the size of the network to compute the full Hessian matrix. We use simply SGD to learn the tasks in a sequential fashion.
Additionally, we evaluate the effective rank of the average Hessian for multiple threshold $\lambda$ values, which better captures the effective dimensionality of the matrix column space. 

\subsection{Perturbation score on Rotated-MNIST}

We repeat the experiment in \cref{robustlocalapprox} for the Rotated-MNIST challenge with the standard network configuration. Similarly to the Split CIFAR-10 setting, we train the network with SGD on the $20$ tasks sequentially. We evaluate the perturbation score for the learning rate which achieves the lowest task loss for all tasks, such that the gradients are approximately $0$. 

We observe two main differences with the Split CIFAR-10 results. First, the curve does not converge to $1$ as we increase the radius, which indicates that the loss landscape of the MLP network, compared to the ResNet18 is more close to convex around the tasks local minima. Second, the absolute magnitude of the perturbation score is significantly lower on the Rotated-MNIST benchmark. We believe this is an effect of the difference in network size ($80$K parameters for Rotated-MNIST and $11$M parameters for Split CIFAR-10). Similarly to the Split-CIFAR10 results, we see a remarkable similarity in the curves shape between tasks, which again suggests that the loss landscape geometry across local minima is similar across tasks.  

\begin{figure}
    \centering
    \includegraphics[width=0.8\linewidth]{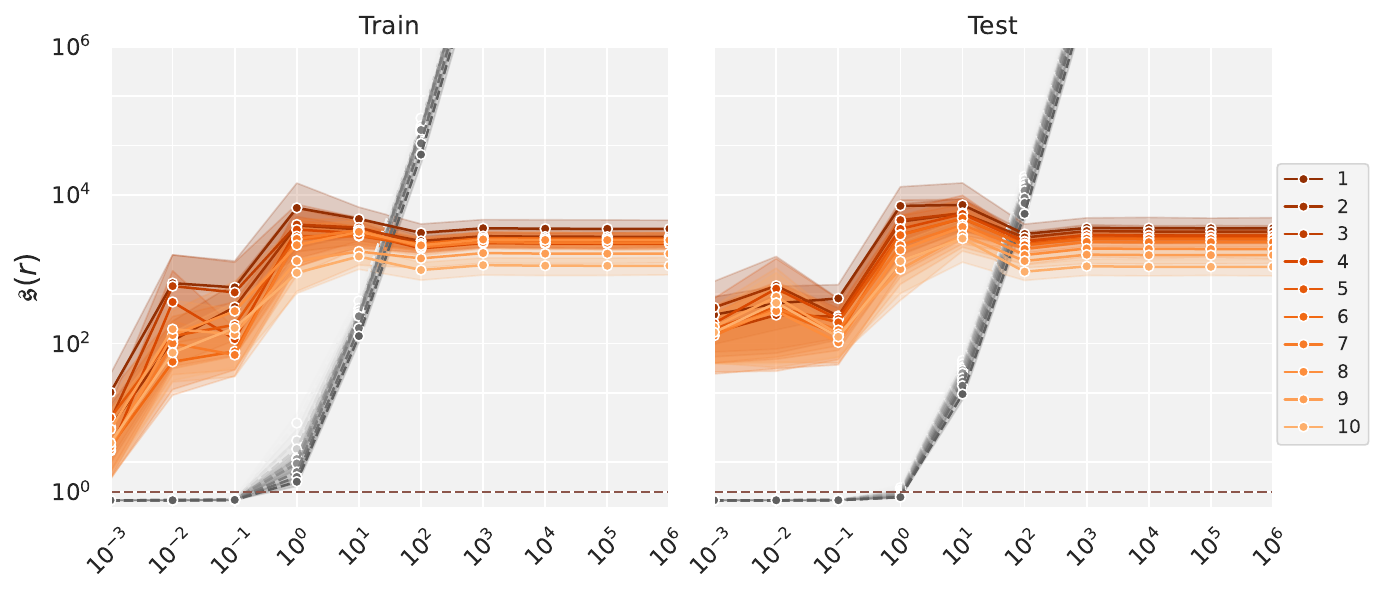}
    \caption{ 
    In \textcolor{orange}{orange}, the perturbation score $\mathfrak{s}(r)$ (\cref{score-ptb}) and in \textcolor{gray}{gray}, the task loss, evaluated on train and test data on the Rotated-MNIST 20 tasks. The shaded area around the curves reflects standard deviation across tasks. Different lines correspond to different perturbation directions (the first $10$ eigenvectors of the the corresponding loss). We evaluate the curves for multiple values of $r$ on a logarithmic scale in the range $[10^{-3}, 10^{6}]$.  The shape of the curve is remarkably stable across tasks.}
    \label{fig:perturbationRM}
\end{figure}

\subsection{Evaluation of Stable SGD}

\begin{figure}
    \centering
    \includegraphics[width=0.8\linewidth]{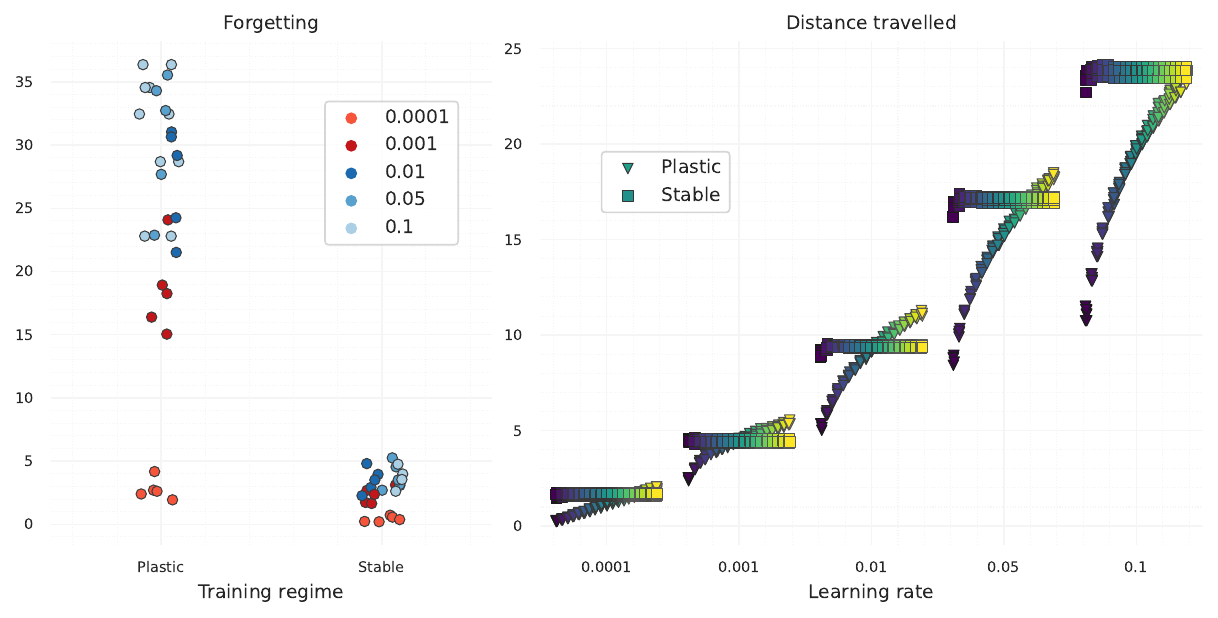}
    \caption{ Comparison of the \emph{plastic} and \emph{stable} versions of SGD in terms of average forgetting and distance travelled in the parameter space.}
    \label{fig:sgdstable}
\end{figure}
Finally, we use our experimental setting to compare SGD under the \emph{Platic} and \emph{Stable} training regimes, studied by \citet{mirzadeh2020understanding}. In the latter case the learning hyperparameters are adjusted to achieve the best tradeoff between plasticity and stability, according to their analysis of the geometry of the task local minima. In particular, stable SGD uses dropout, large initial learning rate and small batch size in order to converge to flatter minima, and limit the distance travelled in the parameter space. We repeat their experiment on Rotated-MNIST, copying their exact hyperparameter setup (except for the number of learning rate and number of epochs, which we increase to $5$). In \cref{fig:sgdstable} we compare the \emph{Plastic} and \emph{Stable} in terms of forgetting and distance traveled in the parameter space for multiple learning rates. We observe a lower forgetting in the stable regime across all learning rates, which suggests that the beneficial properties of the stable configuration are robust to the specific learning rate value.  Additionally, we observe that the trajectory in the parameter space converges quickly to a fixed region, limiting the movement from one task to the next. Overall, our observations confirm the results of \citet{mirzadeh2020understanding}. 

We may argue that SGD is a global algorithm, as its objective coincides with the the current task loss $L_t(\bm\theta)$, thus implicitly using the constant approximation $\hat{L}_i(\bm\theta)=0$ for any $i<t$. However, it displays the behaviour of a local algorithm (\cref{fig:sgdstable}), with higher learning rates yielding higher forgetting in both the stable and plastic regimes. We speculate that this effect might be due to the implicit regularisation of gradient descent \citep{smith2021origin}, which in the linear case can be shown to pull the task solution towards its initialisation \citep{gunasekar2018characterizing}, coinciding with the previous task solution. In this way, we have shown how to use the characterisation of the forgetting patterns of local and global algorithms in order to capture the implicit biases of existing algorithms, such as SGD. 

\end{document}